\documentclass[11pt]{article}

\usepackage[preprint]{acl}
\providecommand{\shortauthors}{}

\usepackage{times}
\usepackage{float}
\usepackage{latexsym}
\usepackage[T1]{fontenc}
\usepackage[utf8]{inputenc}
\usepackage{microtype}
\usepackage{inconsolata}
\usepackage{lineno}
\usepackage{graphicx}
\usepackage{tabularx}
\usepackage[utf8]{inputenc} 
\usepackage[T1]{fontenc}    
\usepackage{longtable, array}
\usepackage{ltablex}
\usepackage{booktabs}
\usepackage{subcaption} 
\usepackage{url}
\usepackage{graphicx}
\usepackage{amssymb}
\usepackage{amsmath}
\usepackage{amsfonts}
\usepackage{url}
\usepackage{bbm}
\usepackage{longtable}
\usepackage{rotating}
\usepackage{multirow}
\usepackage{mathrsfs}
\usepackage{enumitem}
\usepackage{adjustbox}
\usepackage{hyperref}
\usepackage{pgfplots}
\usetikzlibrary{pgfplots.dateplot}
\usepackage{filecontents}
\usepackage[many]{tcolorbox}
\usepackage{xcolor}
\usepackage{listings}
\usepackage{wrapfig}
\usepackage{cuted}

\newcommand{\eg}{\textit{e}.\textit{g}.}

\DeclareMathOperator*{\argmin}{arg\,min}

\begin{document}

\title{Navigating User Behavior toward Personalized Multimodal Generation}

\author{
  Hengji Zhou$^{1*}$, Yufeng Liu$^{1*}$, Ye Liu$^1$, Yong Xu$^1$, 
  Lianghao Xia$^{2\dagger}$, Liqiang Nie$^2$ \\
  $^1$South China University of Technology \\
  $^2$Harbin Institute of Technology, Shenzhen \\
  \texttt{hengjizhou01@gmail.com}, \texttt{202330361751@mail.scut.edu.cn}, \\
  \texttt{202330451251@mail.scut.edu.cn}, \texttt{yxu@scut.edu.cn}, \\
  \texttt{aka\_xia@foxmail.com}, \texttt{nieliqiang@gmail.com}
}

\renewcommand{\shortauthors}{Zhou et al.}
\def\model{NaviGen}
\maketitle
\footnotetext[1]{$^*$Hengji Zhou and Yufeng Liu have equal contribution to this work.}
\footnotetext[2]{$^\dagger$Lianghao Xia is the corresponding author.}

\begin{abstract}
Modern AIGC pipelines deliver high-fidelity images and videos but presuppose a well-formed creation instruction, while end users rarely articulate visual details, leaving generators misaligned with user demand. We study \textbf{personalized content generation}, which turns a user's interaction history into an executable instruction for downstream synthesis, and identify two obstacles: behavior must be encoded in a form legible to language reasoning, and the model must acquire instruction-writing skill absent from both pretraining and behavior data. We propose \textbf{\model}, which represents each item with a dual identifier coupling a collaborative code and a textual code as a behavioral substrate and a semantic bridge in one token stream. On this representation, a two-stage SFT+RL pipeline first distills preference reasoning and instruction writing from evolutionarily searched supervision, then aligns generation with user intent through hierarchical and self-consistent rewards. Experiments across product, game, and short-video domains show that \model\ improves personalized image and video generation, strengthens next-item prediction, and yields more specific, relevant, and visually generatable instructions. Our code is released at: \url{https://github.com/iLearn-Lab/NaviGen}.

\end{abstract}

\section{Introduction}
\label{sec:intro}

Multimodal content generation, such as text-to-image posters and short videos, is rapidly becoming a core productivity layer for media, marketing, and e-commerce~\citep{xu2025personalized, ling2026ragar}. A modern creation pipeline refines a textual instruction with a language model and renders it into an image or video via a text-to-vision generator~\citep{yang2025cogvideox, seedance2026seedance}, where the instruction serves as the central control signal of what is generated and how it looks.

Recent progress along this pipeline broadly falls into three lines. \emph{(i) Generation backbones:} diffusion and autoregressive transformers for text-to-image and text-to-video deliver high-fidelity rendering from textual prompts~\citep{yang2025cogvideox, seedance2026seedance}. \emph{(ii) Instruction enrichment:} LLM-based prompt expansion and multi-agent creation systems turn short user inputs into detailed, structured instructions that better exploit generator capacity~\citep{xu2025personalized, an2026unictokens, dang2026multi}. \emph{(iii) Conditional control:} reference-, layout-, or identity-conditioned generation injects external signals for fine-grained control over the output~\citep{zhao2025local, ling2026ragar}.

\begin{figure}
    \centering
    \includegraphics[width=\columnwidth]{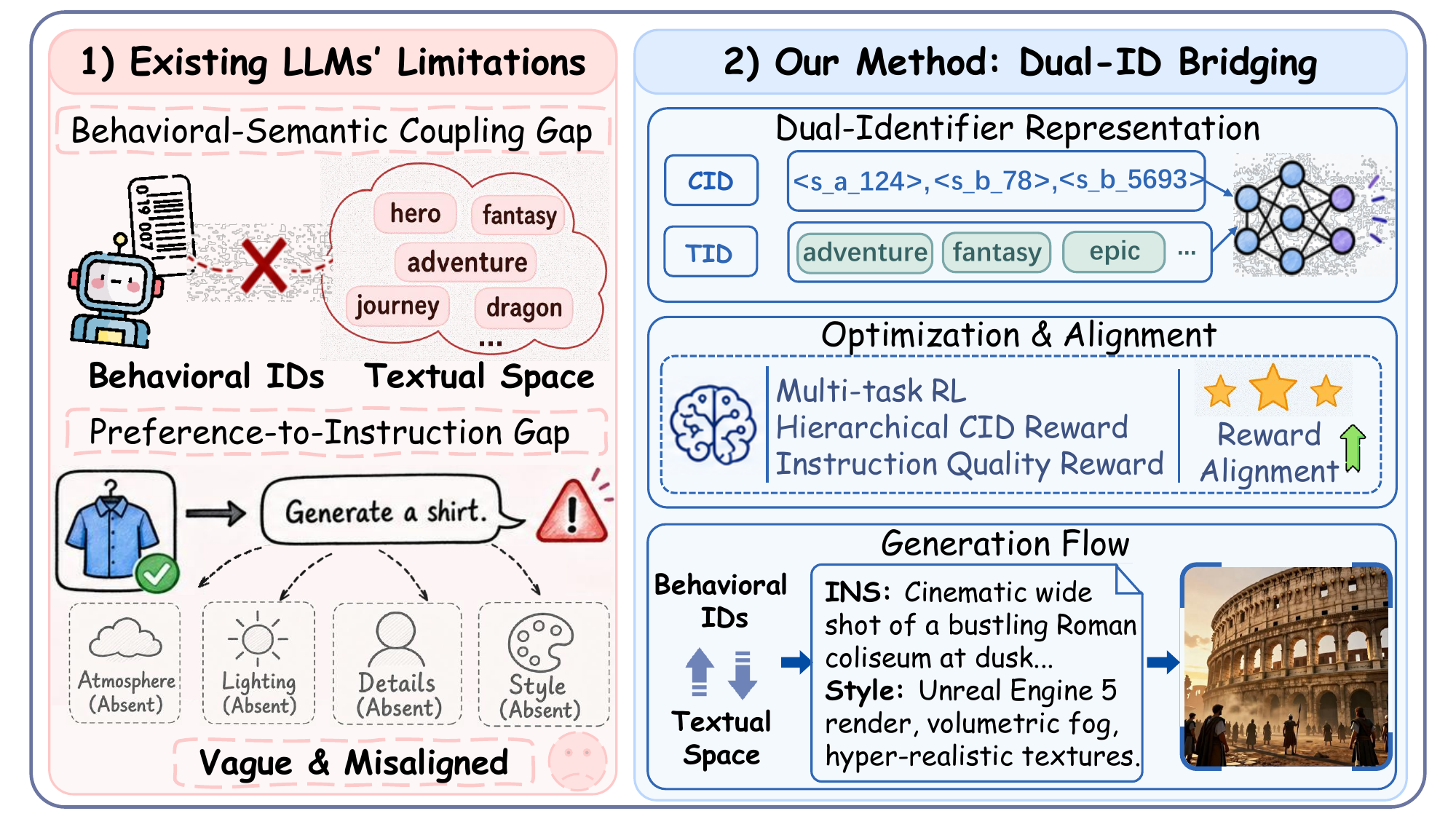}
    \vspace{-0.2in}
    \caption{Personalized multimodal content generation.}
    \label{fig:intro}
    \vspace{-0.15in}
\end{figure}

However, all of these methods presuppose \textbf{a well-formed creation instruction as input}, leaving a fundamental question unanswered: \emph{whose taste is the content actually for?} End consumers differ widely in taste and rarely articulate visual details, yet whether the content resonates with them is what ultimately matters. Without a path from user signals to a concrete instruction, even the strongest generators produce content that is generic or misaligned with real demand. This gap motivates a new problem we term \textbf{\emph{personalized content generation}}: translating a user's implicit behavior history into a creation instruction that steers the downstream generator toward content the user truly wants.

Realizing this paradigm raises two central challenges, on the input and output sides of the language model respectively. \textbf{(C1) Representation gap between behavior and language.} A user's behavior history must be encoded into a form the LM can reason over, yet no single representation suffices: ID-based encodings preserve behavioral structure but remain opaque in the LM's semantic space, while raw textual metadata is expressive but verbose, tempting the model to paraphrase history rather than infer preference. A workable representation must carry behavioral signal and stay legible to language reasoning at once. \textbf{(C2) Capability gap between understanding preference and writing instructions.} Even given such a representation, \emph{knowing what a user likes} and \emph{writing a good creation instruction} are two distinct skills—the latter neither cultivated by language pretraining nor reflected in user behavior data, leaving the model with no natural source to acquire it.

We propose \model, whose two core designs map one-to-one onto these challenges. To close the representation gap (C1), \model\ encodes each item with a \textbf{dual-identifier} scheme: a collaborative identifier (CID) captures its behavioral role via residual vector quantization, while a textual identifier (TID) compresses its textual semantics into ordered, standardized terms. Together they give the language model a compact behavioral substrate and a controllable language bridge in a single token stream. To close the capability gap (C2), \model\ adopts a \textbf{two-stage SFT+RL pipeline}. The SFT stage learns from history-to-instruction supervision synthesized by evolutionary search under an LLM judge, teaching the model to reason about preference evolution rather than paraphrase history. The RL stage jointly optimizes two complementary rewards: a hierarchical CID reward for preference correctness, and a triangular instruction-aware reward defined over the generated instruction, the predicted target semantics, and the ground-truth target semantics, which together drive the model toward generation-ready instructions.

Our contributions are summarized as follows:\vspace{-0.1in}
\begin{itemize}[leftmargin=*]
    \item We propose \model, a unified framework that turns user behavior sequences into generation-ready creation instructions for personalized AIGC, bridging behavioral modeling and controllable content generation in a single pipeline.\vspace{-0.1in}
    
    \item We introduce a dual-identifier representation that couples a residual-quantized CID with an ordered, length-flexible TID, jointly providing a compact behavioral substrate and a controllable semantic bridge within one token stream.
    
    \item We design a two-stage SFT+RL pipeline that requires no human-written instructions: evolutionary search with an LLM judge synthesizes supervision, while GRPO uses a hierarchical CID reward and a triangular instruction-aware reward enforcing closed-loop self-consistency.\vspace{-0.1in}
    
    \item Across product, game, and short-video domains, \model\ consistently improves personalized image and video generation quality, CID-space next-item prediction accuracy, and instruction specificity, relevance, and visual generatability.
\end{itemize}

\section{Preliminary}
\label{sec:preliminary}

\noindent\textbf{Personalized Content Generation}. We consider the setting of consumer-facing personalized AIGC, where a multimodal generative model synthesizes visual content tailored to an individual user. Given a textual creation instruction $\mathcal{I} \in \mathcal{T}$, an off-the-shelf generator $g_{\phi}$ produces the final output:
{\setlength{\abovedisplayskip}{4pt}
\setlength{\belowdisplayskip}{4pt}
\begin{equation}
\mathcal{O} \;=\; g_{\phi}(\mathcal{I}),
\end{equation}}

\noindent where $\mathcal{O} \in \mathcal{Y}$ denotes the generated content in the target modality space (\eg, image or video), and $g_{\phi}$ remains fixed during our training. Under this setting, the quality of personalized generation is fundamentally bottlenecked by the quality of $\mathcal{I}$, while end consumers cannot be expected to author such instructions by hand. This motivates the need for an automatic instruction generator $f_{\theta}$ that produces $\mathcal{I}$ on the user's behalf.

\noindent\textbf{Behavior as Implicit Preference Evidence}. To drive $f_{\theta}$ toward user-specific generation, we leverage the user's observed interaction history as an implicit signal of visual preference. For a given user, we denote this history as an ordered sequence
{\setlength{\abovedisplayskip}{4pt}
\setlength{\belowdisplayskip}{4pt}
\begin{equation}
\mathcal{H}_u \;=\; \langle x_{1}, x_{2}, \ldots, x_{n} \rangle,
\end{equation}}

\noindent where each $x_{k}$ is an item the user has previously engaged with (\eg, clicked, viewed, or purchased), associated with its visual and semantic attributes. We treat $\mathcal{H}_u$ as \emph{preference evidence}: a record from which the user's latent visual taste can be inferred and projected forward into a creative direction.

\noindent\textbf{Task Formulation: Behavior-Conditioned Instruction Generation}. Given a user's history $\mathcal{H}_u$, our goal is to learn an instruction generator $f_{\theta}$ that produces a free-form textual instruction:
{\setlength{\abovedisplayskip}{4pt}
\setlength{\belowdisplayskip}{4pt}
\begin{equation}
\mathcal{I} \;=\; f_{\theta}(\mathcal{H}_u),
\quad
\max_{\theta} \; P_{\theta}\!\left(\mathcal{I} \mid \mathcal{H}_u\right).
\end{equation}}

\noindent We say that an instruction $\mathcal{I}$ is \emph{generation-ready} if it satisfies two properties:
(1) \emph{preference alignment}: $\mathcal{I}$ faithfully captures the user-specific visual preferences evidenced by $\mathcal{H}_u$; and
(2) \emph{generation feasibility}: $\mathcal{I}$ is sufficiently concrete and visually grounded to serve as an effective conditioning signal for the downstream generator $g_{\phi}$.
The objective of this work is to design $f_{\theta}$ such that, for arbitrary user histories, it consistently emits generation-ready instructions, thereby bridging implicit user behavior and high-quality personalized visual synthesis.
\section{Method}
\label{sec:method}

\begin{figure*}[t]
  \centering
  \includegraphics[width=1\linewidth]{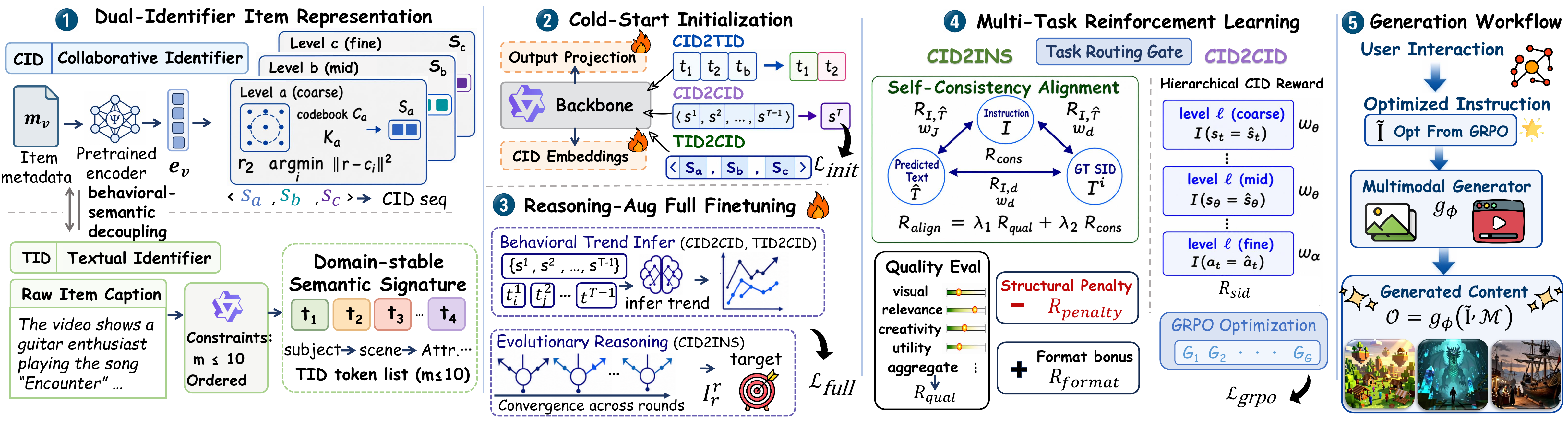}
  \vspace{-0.2in}
  \caption{Overall architecture of the proposed \model\ framework for personalized multimodal generation.} 
  \label{fig:framework}
  \vspace{-0.12in}
\end{figure*}

This section presents the technical details of \model, with overall architecture shown in Figure~\ref{fig:framework}.\vspace{-0.05in}

\subsection{Dual-Identifier Behavior Encoding}

To make user behavior $\mathcal{H}_u$ legible to an LLM, each entry $x_k \in \mathcal{H}_u$ must be serialized into tokens within the model's vocabulary. A naive choice, directly feeding captions or metadata, is verbose and injects redundant noise that slows optimization and blurs preference signals. We therefore encode each entry with a compact \emph{dual identifier}, decoupling sequence-level identity from semantic grounding so that neither role compromises the other.

\noindent\textbf{Collaborative Identifier (CID).} Inspired by LLMs for collaborative filtering~\citep{onerec}, the CID encodes an entry's role within user interaction sequences, distilling collaborative patterns observed across the consumer-content interaction graph. Its metadata $m_v$ is first mapped to a continuous embedding $\mathbf{e}_v = \psi(m_v)$ via a pretrained embedding model $\psi$, then quantized through a multi-layer residual K-means process:
{\setlength{\abovedisplayskip}{8pt}
\setlength{\belowdisplayskip}{4pt}
\begin{align}
    s_\ell &= \argmin_k \|\mathbf{r}_\ell - \mathbf{c}_\ell^{k}\|^2, \;\mathbf{r}_{\ell+1} = \mathbf{r}_\ell - \mathbf{c}_\ell^{s_\ell},
\end{align}}

\noindent where $\mathbf{r}_1 = \mathbf{e}_v$, $\mathbf{c}_\ell^{k}$ denotes the $k$-th centroid in the $\ell$-th codebook, and $s_\ell \in \{1, \ldots, K_{\text{cb}}\}$ is the discrete code assigned to level $\ell$. The resulting CID is a three-level residual token sequence:
{\setlength{\abovedisplayskip}{6pt}
\setlength{\belowdisplayskip}{6pt}
\begin{equation}
    \text{cid}(v) = \langle\; s_1(v),\; s_2(v),\; s_3(v)\;\rangle.
\end{equation}}
This hierarchy enables multi-granularity modeling and partial-credit supervision, as matching any level yields a meaningful signal. Each CID token is added to the vocabulary and initialized via dedicated embedding training (Section~\ref{subsec:sft-stage1}).

\noindent\textbf{Textual Identifier (TID).} Unlike existing work that treats TIDs as fixed-length targets for next-item prediction~\citep{zhang2026unleashing}, we note that semantically equivalent textual variants may correspond to different terms, making exact next-TID prediction overly restrictive; meanwhile, entries vary in semantic complexity and thus require different numbers of terms. We therefore construct variable-length TIDs by imposing only an upper bound on the number of terms:
{\setlength{\abovedisplayskip}{8pt}
\setlength{\belowdisplayskip}{8pt}
\begin{equation}
    \text{tid}(v) = [t_1, t_2, \ldots, t_m], \quad m \leq 10,
\end{equation}}
where each $t_k$ is a concise phrase capturing a core semantic dimension (\eg, subject category, key attribute). Terms are ordered by importance and produced by compressing the original caption through an LLM with controlled output constraints. Unlike free-form text, the TID provides a compact, deduplicated, and domain-stable semantic signature that serves as the bridge between sequence-level preference modeling and instruction generation.\vspace{-0.03in}

\subsection{Reasoning-Infused Supervised Tuning}
\label{subsec:sft}

\model\ employs two-stage supervised fine-tuning that progressively builds from identifier embeddings to full reasoning-capable generation.

\subsubsection{Cold-Start Embedding Initialization}
\label{subsec:sft-stage1}

To prevent randomly initialized CID embeddings from destabilizing pretrained weights through noisy gradients, we decouple representation acquisition from backbone adaptation: all pretrained weights are frozen, and only parameters tied to the new tokens are updated. Let $\mathcal{D}_{\text{init}}$ denote the auxiliary training set with token sequences $\mathbf{s}_i$ and target tokens $y_i$. The embedding initialization loss is:
\begin{equation}
    \mathcal{L}_{\text{init}} = -\sum_{i} \sum_{t} \log p(y_{i,t} \mid \mathbf{s}_{i,<t}; \mathcal{E}_{\text{CID}}, \mathbf{W}_{\text{out}})
\end{equation}
where $\mathcal{E}_{\text{CID}}$ and $\mathbf{W}_{\text{out}}$ are the learnable CID embeddings and output projection layer. Three auxiliary tasks establish bidirectional CID-TID alignment:

\noindent\textbf{CID2TID.} Mapping CID to its corresponding TID, grounding behavioral codes in semantic terms.

\noindent\textbf{TID2CID.} Inverse TID-to-CID mapping, constructing behavioral identifiers from semantic signals.

\noindent\textbf{CID2CID.} Predicting a target CID from a history of CIDs, capturing sequential behavioral patterns.

\subsubsection{Reasoning-Augmented Full Finetuning}
\label{subsec:sft-stage2}

The initialization stage equips the model with stable CIDs, but two capabilities essential for behavior-conditioned instruction generation remain absent: (i) translating inferred preferences into generation-ready instructions, and (ii) reasoning over how user interests evolve along the interaction history. To instill both, we unfreeze all parameters and augment the existing objectives with a new \textbf{CID2INS} task and chain-of-thought supervision.

\noindent\textbf{CID2INS.} AIGC instructions are synthesized through an evolution-inspired search that progressively refines candidate instructions toward user-aligned visual semantics. Starting from three founder strategies (conservative, balanced, exploratory), each round selects the two strongest candidates via a multi-dimensional judge and produces two offspring through crossover and controlled mutation. Given the candidate trajectory $\{\mathcal{I}^{(r)}\}_{r=0}^{R}$, the final instruction is:
{\setlength{\abovedisplayskip}{6pt}
\setlength{\belowdisplayskip}{6pt}
\begin{equation}
    \mathcal{I}^{\star} = \operatorname*{arg\,max}_{\mathcal{I} \in \mathcal{P}_{\text{f}}} \; \mathcal{S}_{\text{judge}}(\mathcal{I}),
\end{equation}}
where $\mathcal{P}_{\text{f}}$ is the final population and $\mathcal{S}_{\text{judge}}$ is the LLM-based multi-dimensional scorer. The model is jointly supervised to predict the target TID, anchoring the generated instruction to the intended visual semantics and preventing semantic drift.

\noindent\textbf{Reasoning.} Chain-of-thought traces are distilled from a teacher model. For \textbf{CID2CID}, the teacher articulates how user preferences evolve along the historical identifier sequence without explicitly referencing the target item. Formally, given an identifier sequence $h_{1:n} = (h_1, \ldots, h_n)$ where $h_i$ denotes the CID at step $i$, the teacher distills consecutive preference shifts into a reasoning chain:
\begin{equation}
\mathcal{T}\big( \{h_{i-1} \rightarrow h_i\}_{i=2}^{n} \big),
\end{equation}
where $\mathcal{T}(\cdot)$ aggregates the evolving preference trajectory across the interaction history. For \textbf{CID2INS}, the reasoning summarizes the full evolutionary trajectory---how candidate instructions converged toward the target semantics across rounds. Formally, starting from founder strategies that form the initial population $\mathcal{P}^{(0)}$, each round $r \in \{1,\dots,R\}$ selects the two highest-scoring candidates under $\mathcal{S}_{\text{judge}}$ and updates the population via elitism, crossover, and mutation:
\begin{equation}
\mathcal{P}^{(r)} = \bigl\{\, \mathcal{I}^{(r)}_{\text{elite}},\; \mathcal{I}^{(r)}_{\text{elite}} \oplus \mathcal{I}^{(r)}_{\text{mate}},\; \mu(\mathcal{I}^{(r)}_{\text{elite}}) \,\bigr\},
\end{equation}
where $\mathcal{I}^{(r)}_{\text{elite}}$ and $\mathcal{I}^{(r)}_{\text{mate}}$ are the top-2 candidates in $\mathcal{P}^{(r-1)}$, $\oplus$ fuses the elite's target-facing anchor with the mate's visual expressiveness, and $\mu(\cdot)$ applies a controlled mutation that preserves the semantic core. The resulting score trajectory across rounds teaches the model to connect evolutionary search dynamics with the final output.

Let $\mathcal{D}_{\text{full}}$ denote the reasoning-augmented training set with input sequences $\mathbf{s}_i$ and output sequences $\mathbf{y}_i$ containing reasoning traces and structured answers. The full finetuning objective is:
\begin{equation}
    \mathcal{L}_{\text{full}} = -\sum_{i} \sum_{t} \log p(\mathbf{y}_{i,t} \mid \mathbf{s}_{i,<t}; \theta),
\end{equation}
where $p(\cdot \mid \mathbf{s}_{i,<t}; \theta)$ denotes the token-level probability under all trainable parameters $\theta$. The resulting model jointly performs preference-grounded reasoning and behavior-conditioned instruction generation for downstream synthesis.\vspace{-0.05in}

\subsection{Multi-Task Reinforcement Learning}
\label{subsec:grpo}

Supervised fine-tuning produces a competent generator but does not directly optimize for the quality of personalized outputs. \model\ applies GRPO to refine the policy $\pi_\theta$ under task-specific reward signals that better reflect downstream objectives.

\noindent\textbf{Hierarchical CID Reward.} The CID encodes item behavior through three residual levels of decreasing granularity. A match at any level provides a meaningful signal, but coarse agreement is weighted more heavily than fine precision, reflecting the intuition that predicting the right item family matters more than nailing the exact variant. Formally, given a predicted CID $\hat{s} = (\hat{s}_a, \hat{s}_b, \hat{s}_c)$ and ground truth $\tilde{s} = (\tilde{s}_a, \tilde{s}_b, \tilde{s}_c)$, the hierarchical reward is:
\begin{equation}
    R_{\text{cid}} = \sum_{\tau \in \{a,b,c\}} w_\tau \cdot \mathbb{I}[\hat{s}_\tau = \tilde{s}_\tau],
\end{equation}
where $w_a$, $w_b$ and $w_c$ enforce the coarse-to-fine weighting. A small bonus rewards predictions that remain within the valid CID vocabulary even when they do not match the ground truth, encouraging the model to stay within the learned identifier space.

\noindent\textbf{Instruction-Aware Reward.} For \textbf{CID2INS}, the reward combines instruction quality assessment with a triangular self-consistency check. An LLM-based judge evaluates the instruction along four dimensions---\textit{specificity}, \textit{creativity}, \textit{content quality}, and \textit{visual generatability}, aggregated into $R_{\text{qual}}$. Beyond standalone quality, a closed-loop alignment enforces three mutually reinforcing signals: the instruction must anchor to target semantics $R_{\text{ins}{\leftrightarrow}\tilde{t}}$, remain self-consistent with its own prediction $R_{\text{ins}{\leftrightarrow}\hat{t}}$, and the prediction itself must align with ground truth $R_{\hat{t}{\leftrightarrow}\tilde{t}}$. The combined reward is:
\begin{align}
    R_{\text{align}} &= \gamma_1 R_{\text{ins}{\leftrightarrow}\tilde{t}} + \gamma_2 R_{\text{ins}{\leftrightarrow}\hat{t}} + \gamma_3 R_{\hat{t}{\leftrightarrow}\tilde{t}}, \\
    R_{\text{ins}} &= \lambda_1 \cdot R_{\text{qual}} + \lambda_2 \cdot R_{\text{align}},
\end{align}
where $\gamma_1, \gamma_2, \gamma_3$ balance the three alignment signals, and $\lambda_1, \lambda_2$ weight quality against consistency.

\noindent\textbf{Optimization Objective.}
We optimize the GRPO objective over a group of $G$ completions:
\begin{equation}
    \label{eq:grpo-objective}
    \mathcal{J}_{\text{grpo}} = \mathbb{E}_{\mathbf{q}}[\, \mathcal{L}_{\text{clip}} - \kappa \, \mathbb{D}_{\text{KL}}(\pi_\theta \| \pi_{\text{ref}}) \,],
\end{equation}
where $\mathcal{L}_{\text{clip}}$ is the clipped surrogate loss averaged over the group, $\rho_i$ is the importance sampling ratio, $\epsilon$ the clipping range, and $\kappa$ the KL penalty weight. The group-relative advantage $\hat{A}_i$ is driven by a composite reward:
\begin{align}
    R_{\text{task}} =
    \begin{cases}
        w_{\text{cid}} R_{\text{cid}} + R_{\text{bonus}}, & \\[4pt]
        w_{\text{ins}} R_{\text{ins}} + R_{\text{format}} - R_{\text{penalty}}, &
    \end{cases}
\end{align}
where $R_{\text{bonus}}$ encourages vocabulary-range adherence, $R_{\text{format}}$ verifies JSON parseability and reasoning completeness, and $R_{\text{penalty}}$ penalizes structural violations. The optimized instruction $\tilde{\mathcal{I}}$ from this stage then serves as the control signal for the multimodal generator $g_\phi$ defined in Section~\ref{sec:preliminary}.

\section{Evaluation}
\begin{table}[t]
    \small
    \centering
    \caption{Statistics of the experimental datasets.}
    \label{tab:datasets}
    \setlength{\tabcolsep}{1.2mm}
    \vspace{-0.12in}
    \begin{tabular}{ccccc}
      \toprule
      Dataset & \#Users & \#Items & \#Interactions & Density \\
      \midrule
      Product & 30161 & 74487 & 161405 & 7.18e-5 \\
      Games & 25912 & 35443 & 144337 & 1.57e-4 \\
      Short Videos & 35259 & 153333 & 209172 & 3.87e-5 \\
      \bottomrule
    \end{tabular}
    \vspace{-0.15in}
\end{table}

We evaluate \model\ by answering five research questions. \textbf{RQ1} compares personalized AIGC generation performance. \textbf{RQ2} studies the contribution of key modules. \textbf{RQ3} examines collaborative identifier prediction in the CID space. \textbf{RQ4} analyzes the effect of SFT and RL steps. \textbf{RQ5} examines qualitative instruction cases.

\begin{table*}[t]
    \small
    \centering
    \caption{Overall performance comparison on personalized AIGC instruction generation. Aesthetic and Novelty are averaged over three runs. Excluding Oracle, boldface marks the best image-generation result, while boldface with superscript $\star$ marks the best video-generation result.}
    \label{tab:overall-performance}
    \setlength{\tabcolsep}{1.3pt}
    \renewcommand{\arraystretch}{1.2}
    \vspace{-0.12in}
    \begin{tabular}{cc|cccc|cccc|cccc}
      \hline
      \multirow{2}{*}{Method} & \multirow{2}{*}{Output}
      & \multicolumn{4}{c|}{Games}
      & \multicolumn{4}{c|}{Product}
      & \multicolumn{4}{c}{Short Videos} \\
      \cline{3-14}
      & & Cons. & Nov. & Aes. & Rel.
        & Cons. & Nov. & Aes. & Rel.
        & Cons. & Nov. & Aes. & Rel. \\
      \hline
      \multirow{1}{*}{PMG}
        & Image & 0.3281 & 0.5084 & 0.7424 & 0.4053 & \textbf{0.3282} & 0.5091 & 0.7677 & 0.4044 & 0.3316 & 0.5725 & 0.7376 & 0.3249 \\
      \multirow{1}{*}{PIGEON} & Image & 0.3121 & 0.5449 & 0.7352 & 0.4382 & 0.3104 & 0.4662 & 0.7086 & 0.4347 & 0.3150 & 0.5510 & 0.7291 & 0.3571 \\
      \multirow{1}{*}{RAGAR} & Image & 0.3148 & 0.5002 & 0.7696 & 0.4475 & 0.3122 & 0.4223 & 0.7171 & 0.4204 & 0.3195 & 0.5713 & 0.7410 & 0.3330 \\
      \hline
      \hline
      \multirow{2}{*}{Cipher}
        & Image & 0.3213 & 0.5403 & 0.7501 & 0.4273 & 0.3193 & 0.4595 & 0.7362 & 0.4038 & 0.3265 & 0.5936 & 0.7311 & 0.3963 \\
      \cline{2-14}
        & Video & 0.6315 & 0.5245 & 0.5855 & 0.4307 & 0.6104 & 0.4020 & 0.4423 & 0.3890 & 0.6379 & \textbf{0.6280}$^{\star}$ & \textbf{0.6830}$^{\star}$ & 0.3595 \\
        \hline
      \multirow{2}{*}{Prose}
        & Image & 0.3173 & 0.5796 & 0.7804 & 0.4052 & 0.3130 & 0.4958 & 0.7398 & 0.4298 & 0.3202 & 0.5936 & 0.7475 & 0.3034 \\
      \cline{2-14}
        & Video & 0.6325 & 0.4740 & 0.5215 & \textbf{0.4560}$^{\star}$ & 0.6282 & 0.4640 & 0.4185 & 0.4199 & 0.6459 & 0.5355 & 0.6245 & 0.3554 \\
        \hline
      \multirow{2}{*}{TRIPLE}
        & Image & 0.3152 & 0.5082 & 0.7472 & 0.4273 & 0.3093 & 0.4273 & 0.7182 & 0.4038 & 0.3164 & 0.5654 & 0.7236 & 0.3963 \\
      \cline{2-14}
        & Video & 0.6203 & 0.4170 & 0.4290 & 0.4559 & 0.6217 & 0.3510 & 0.3905 & 0.4093 & 0.6419 & 0.4305 & 0.5330 & 0.3315 \\
        \hline
        \hline
      \multirow{2}{*}{NPC}
        & Image & 0.3238 & 0.5535 & 0.7797 & 0.3728 & 0.3188 & 0.4564 & 0.7338 & 0.3699 & 0.3234 & 0.5825 & 0.7448 & 0.2866 \\
      \cline{2-14}
        & Video & 0.5643 & 0.4935 & 0.4890 & 0.3936 & 0.6006 & 0.3615 & 0.4080 & 0.4012 & 0.6079 & 0.5590 & 0.6335 & 0.3436 \\
      \hline
      \multirow{2}{*}{Oracle}
        & Image & 0.3441 & 0.6830 & 0.8448 & 0.6034 & 0.3296 & 0.6114 & 0.8341 & 0.6143 & 0.3422 & 0.6785 & 0.8155 & 0.5396 \\
      \cline{2-14}
        & Video & 0.6473 & 0.5440 & 0.6275 & 0.5532 & 0.6355 & 0.4460 & 0.5000 & 0.5647 & 0.6510 & 0.5790 & 0.6800 & 0.5248 \\
      \hline
      \hline
      \multirow{2}{*}{\model}
        & Image & \textbf{0.3346} & \textbf{0.6479} & \textbf{0.8423} & \textbf{0.4960} & 0.3173 & \textbf{0.5879} & \textbf{0.8350} & \textbf{0.4839} & \textbf{0.3390} & \textbf{0.6686} & \textbf{0.8272} & \textbf{0.4180} \\
      \cline{2-14}
        & Video & \textbf{0.6449}$^{\star}$ & \textbf{0.5343}$^{\star}$ & \textbf{0.5997}$^{\star}$ & 0.4322 & \textbf{0.6336}$^{\star}$ & \textbf{0.4718}$^{\star}$ & \textbf{0.4468}$^{\star}$ & \textbf{0.4834}$^{\star}$ & \textbf{0.6478}$^{\star}$ & 0.5528 & 0.6636 & \textbf{0.4002}$^{\star}$ \\
      \hline
    \end{tabular}
    \vspace{-0.15in}
\end{table*}

\subsection{Experimental Settings}
\label{sec:exp_setting}
\subsubsection{Datasets and Evaluation Protocols}
Table~\ref{tab:datasets} reports statistics of the three datasets: \textbf{Product} and \textbf{Games} are Amazon review domains with captions derived from product metadata, while \textbf{Short Videos} comes from the OpenOneRec short-video benchmark, where each item has a textual caption~\citep{zhou2025openonerec}. Each data instance is formed by pairing a user's history interaction sequence with a target item, and all instances are split into training, validation, and test sets with an 8:1:1 ratio; all methods are evaluated by Recall@K/NDCG@K. For personalized generation, we sample 1,000 image-generation and 100 video-generation cases from the test split of each dataset, while keeping the AIGC instruction modality consistent across methods. We compare \model\ and baselines along four dimensions: \textbf{Consistency}, measured by image-instruction CLIPScore and, for videos, average CLIPScore over one frame per second ~\citep{hessel2021clipscore}; \textbf{Relevance}, measured by cosine similarity between the generated instruction and the equal-weight embedding of history and target item captions; \textbf{Aesthetic}, where a VLM judge scores visual quality in $[0,1]$ ~\citep{kao2017deep}; and \textbf{Novelty}, where the same judge scores novelty and interestingness in $[0,1]$ ~\citep{vargas2011rank}. Details are provided in Appendix \ref{sec:prompt_temp}

\subsubsection{Baseline Methods}
\model\ is compared with a comprehensive set of baselines, including \textbf{i) Personalized Generation Methods:} PMG~\citep{shen2024pmg}, Pigeon~\citep{xu2025personalized}, RAGAR~\citep{ling2026ragar}, CIPHER~\citep{gao2024aligning}, PROSE~\citep{prose}, TRIPLE~\citep{noh2026triple}; \textbf{ii) Collaborative Filtering Methods:} SASRec~\citep{kang2018self}, TIGER~\citep{rajput2023recommender}, LC-Rec~\citep{lcrec}, OpenOneRec-Pretrain~\citep{zhou2025openonerec}; and \textbf{iii) Prompting and Reference Baselines:} \textbf{NPC} removes user-specific evidence and relies only on a generic task prompt, while \textbf{Oracle} conditions on the ground-truth target caption or TID as a non-deployable target-conditioned reference.

\begin{figure}[t]
  \centering
  \includegraphics[width=\linewidth]{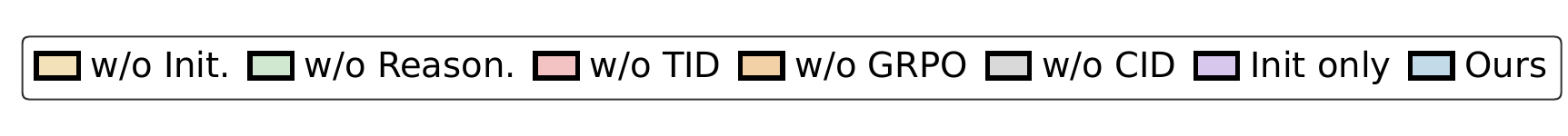}
  \vspace{-0.15in}
  \begin{subfigure}[b]{\linewidth}
    \centering
    \includegraphics[width=0.48\linewidth]{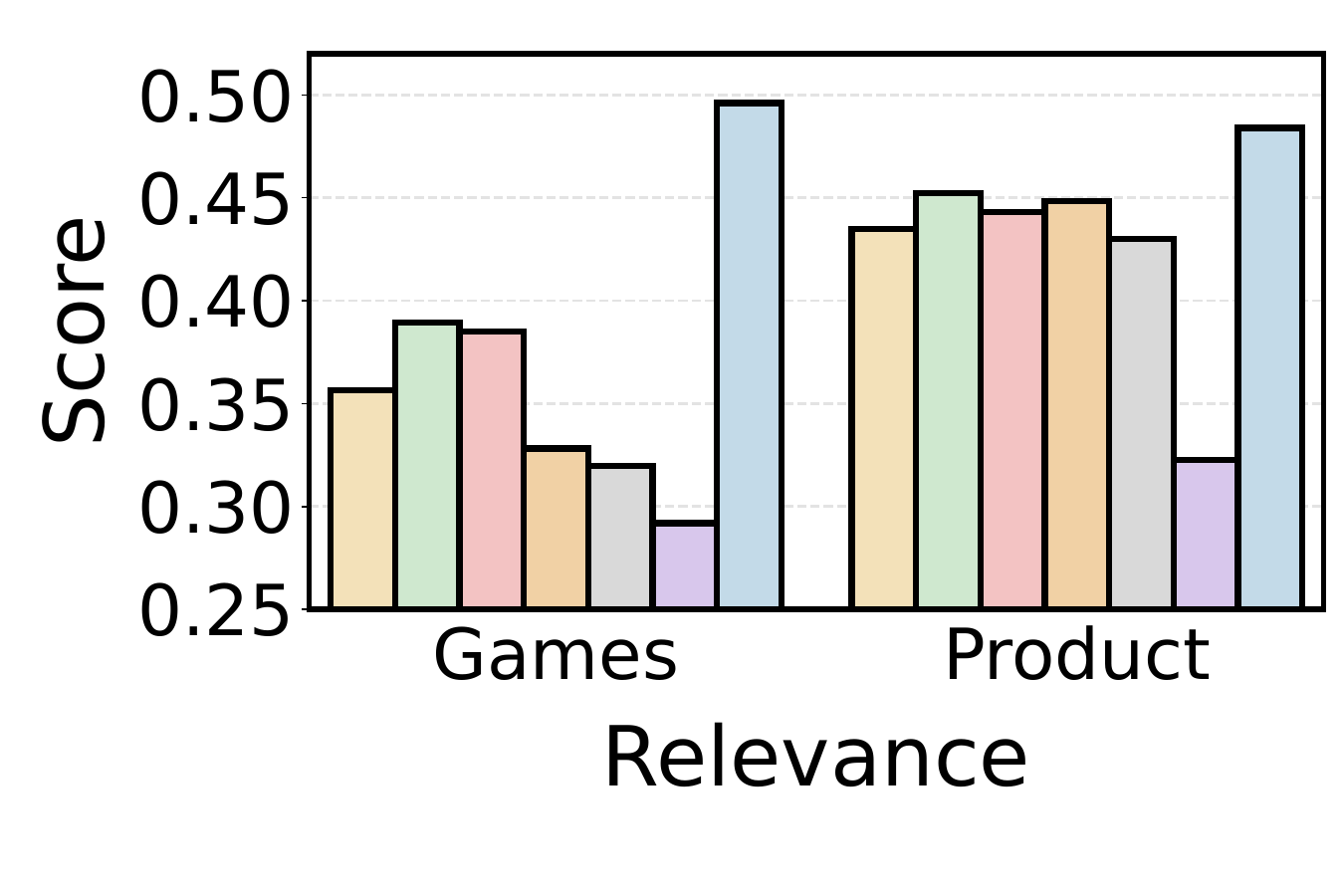}
    \hspace{0.01\linewidth}
    \includegraphics[width=0.48\linewidth]{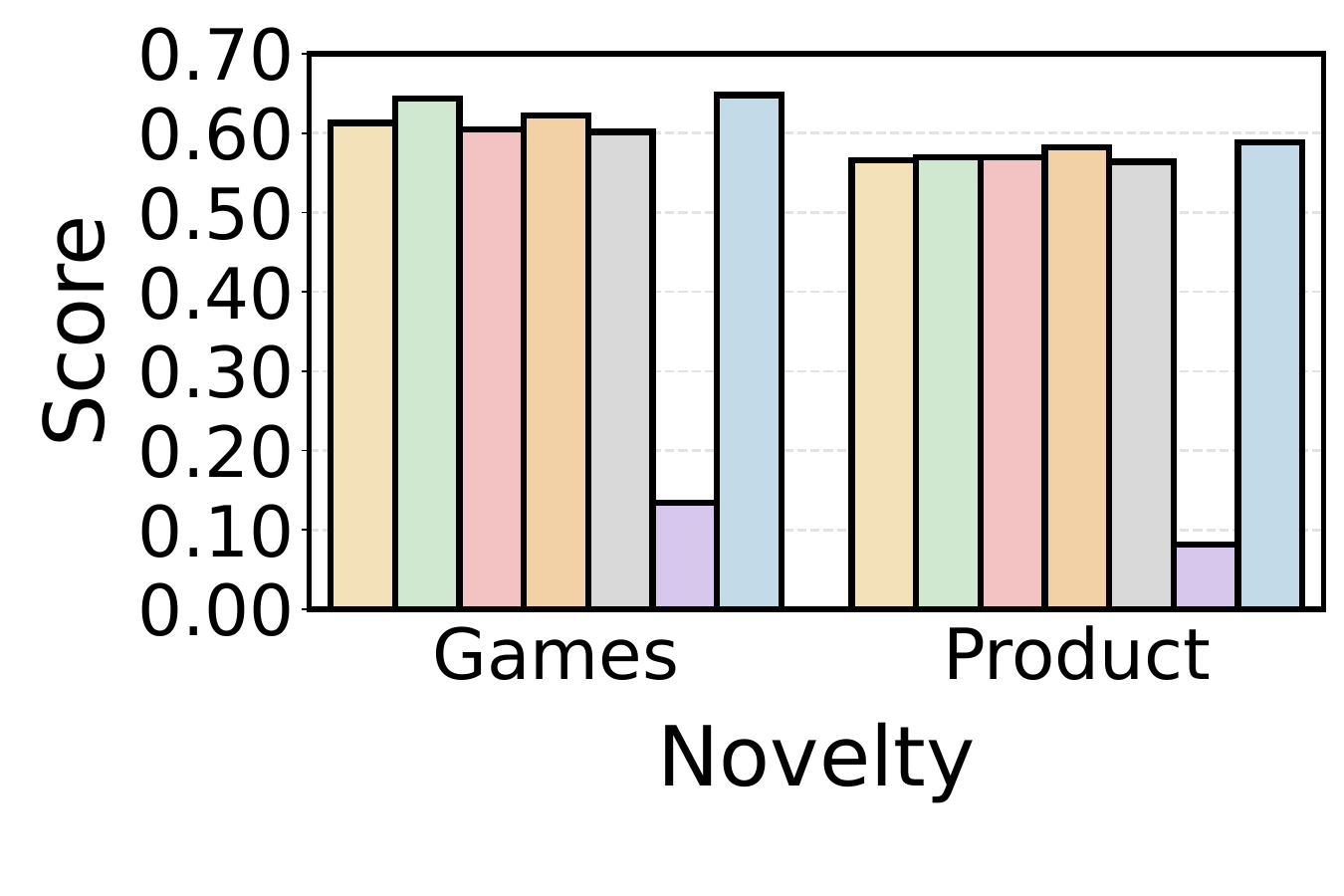}
    \vspace{-0.3in}
    \caption{Image Generation}
    \label{fig:ablation-image-generation}
  \end{subfigure}
  \\[0.15in]
  \begin{subfigure}[b]{\linewidth}
    \centering
    \includegraphics[width=0.48\linewidth]{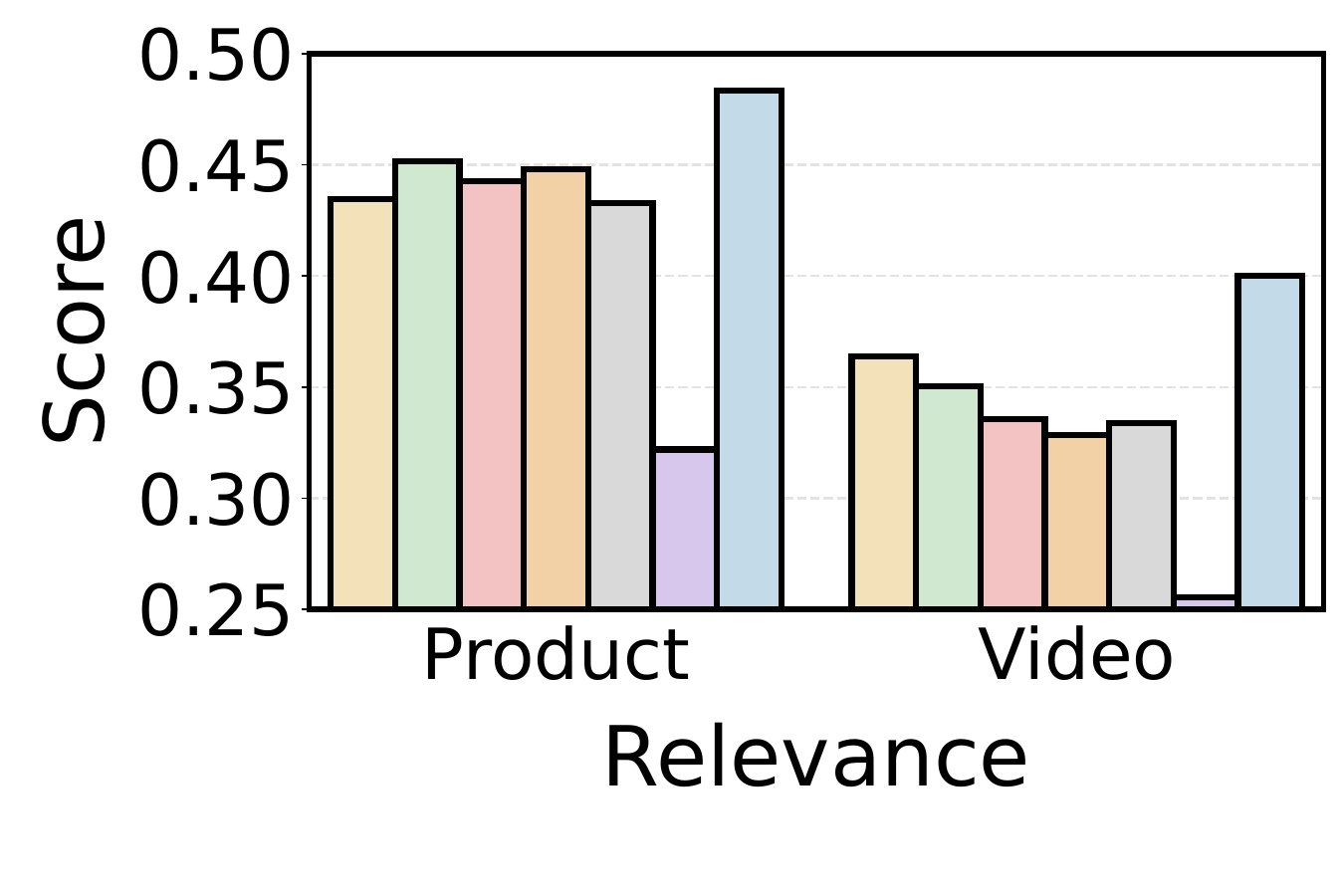}
    \hspace{0.01\linewidth}
    \includegraphics[width=0.48\linewidth]{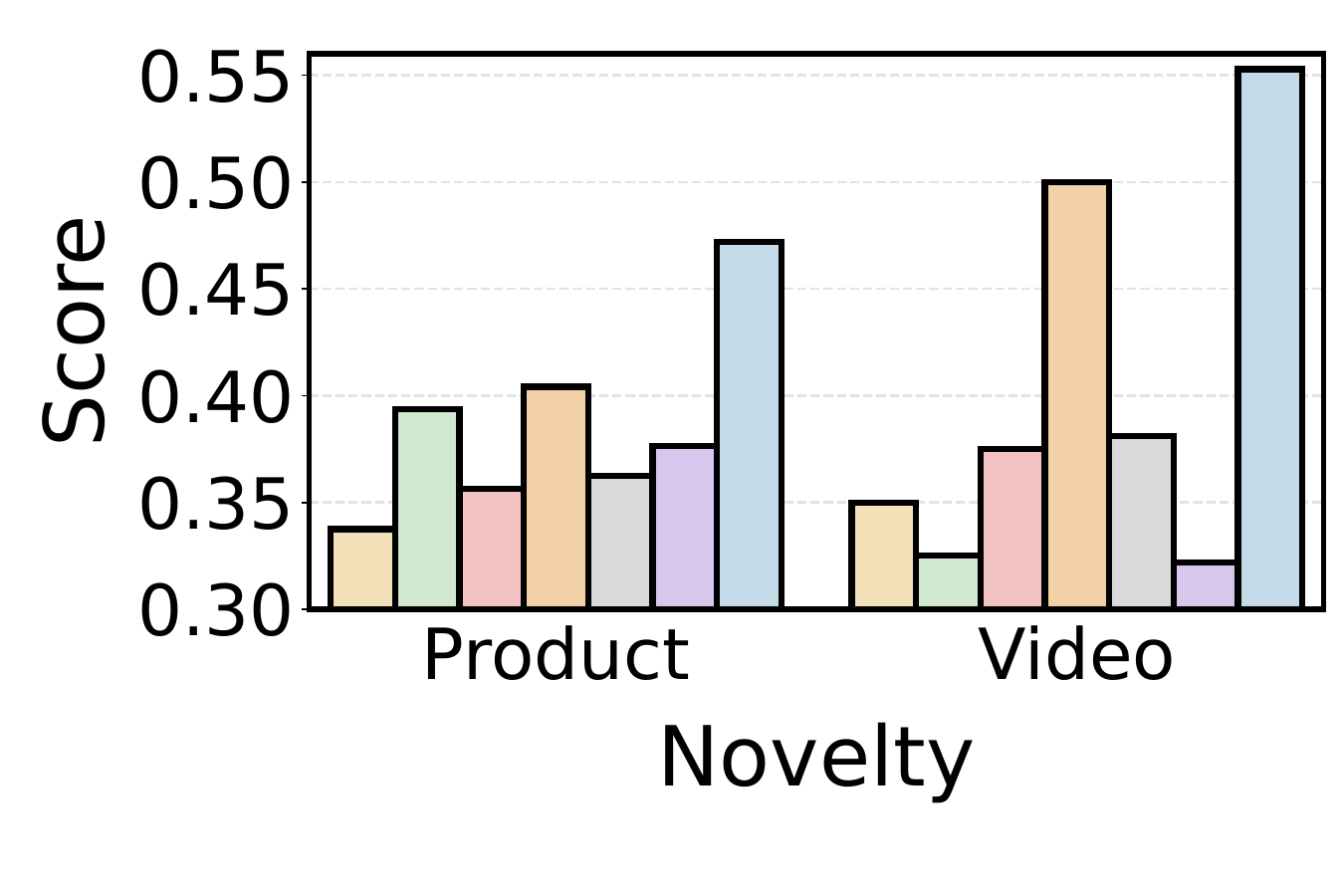}
    \vspace{-0.3in}
    \caption{Video Generation}
    \label{fig:ablation-video-generation}
  \end{subfigure}
  \vspace{-0.25in}
  \caption{Ablation study on image and video generation.}
  \label{fig:ablation-image}
  \vspace{-0.2in}
\end{figure}

\subsubsection{Implementation Details}
All baselines are implemented following their original papers. \model\ uses Qwen3-1.7B as the base language model. Unless otherwise specified, we use Qwen3.5-Flash for auxiliary method components, including Instruction-Aware Reward, TID, and reasoning generation. For both cold-start embedding initialization and full-parameter finetuning, we set the learning rate to $5 \times 10^{-4}$, the warmup ratio to $0.03$, and train for 3 epochs, using AdamW optimization with a cosine learning-rate schedule, a maximum sequence length of 2048, and packed SFT examples. For reinforcement learning, we set the group size to 8, the maximum sequence length to 2048, the weight decay to 0.01, the learning rate to $3 \times 10^{-4}$, the number of training steps to 600, and the batch size to 480. We employ ViT-B/32 for image encoding and text-embedding-v4 for text encoding. We use GLM-5V-Turbo as the VLM judge, Z-Image-Turbo for 512$\times$512 image generation, and Open-Sora 1.3 for 720p video generation at 24 fps with a total of 81 frames. \vspace{-0.05in}

\begin{table*}[t]
    \small
    \centering
    \caption{Performance comparison on collaborative identifier prediction. Best results are highlighted in bold.}
    \label{tab:cid-modeling}
    \setlength{\tabcolsep}{4.7pt}
    \renewcommand{\arraystretch}{1.1}
    \vspace{-0.12in}
    \begin{tabular}{c|cccc|cccc|cccc}
      \hline
      \multirow{2}{*}{Model}
        & \multicolumn{4}{c|}{Games}
        & \multicolumn{4}{c|}{Product}
        & \multicolumn{4}{c}{Short Videos} \\
      \cline{2-13}
        & R@10 & R@20 & N@10 & N@20
        & R@10 & R@20 & N@10 & N@20
        & R@10 & R@20 & N@10 & N@20 \\
      \hline
      SASRec & 0.0281 & 0.0364 & 0.0150 & 0.0171 & 0.0017 & 0.0027 & 0.0006 & 0.0009 & 0.0022 & 0.0039 & 0.0013 & 0.0017 \\
      TIGER & 0.0294 & 0.0446 & 0.0150 & 0.0190 & 0.0051 & 0.0109 & 0.0022 & 0.0037 & 0.0025 & 0.0050 & 0.0014 & 0.0021 \\
      LC-Rec & 0.0463 & 0.0600 & 0.0258 & 0.0289 & 0.0072 & 0.0143 & 0.0040 & 0.0057 & 0.0042 & 0.0080 & 0.0029 & 0.0041 \\
      OneRec & \textbf{0.0471} & 0.0605 & \textbf{0.0267} & 0.0288 & 0.0068 & 0.0137 & 0.0033 & 0.0050 & 0.0047 & 0.0111 & 0.0034 & 0.0053 \\
      \hline
      \model & 0.0431 & \textbf{0.0621} & 0.0247 & \textbf{0.0294} & \textbf{0.0126} & \textbf{0.0211} & \textbf{0.0066} & \textbf{0.0088} & \textbf{0.0155} & \textbf{0.0239} & \textbf{0.0070} & \textbf{0.0092} \\
      \hline
    \end{tabular}
    \vspace{-0.15in}
\end{table*}

\subsection{Overall Performance Comparison (RQ1)}

We compare \model\ against representative baselines on personalized image and video generation; Table~\ref{tab:overall-performance} reports the results, with Oracle as a non-deployable target-conditioned reference. \textbf{Consistent Image-Level Gains.} \model\ achieves the best image-generation performance on most non-oracle comparisons, leading across all metrics on Games and Short Videos and across Novelty, Aesthetic, and Relevance on Product. The only exception is Product Consistency, suggesting that \model\ mainly improves personalized target alignment, creative specificity, and visual quality while keeping image-instruction consistency competitive. \textbf{Video-Level Transfer and Trade-offs.} \model\ obtains the best video-generation result in 9 of 12 non-oracle comparisons, indicating that video generation involves a trade-off among frame-level consistency, novelty, visual quality, and preference-specific relevance. \textbf{Oracle Reference.} Oracle attains the highest Relevance by directly conditioning on target semantics, yet \model\ surpasses it on image Aesthetic for Product and Short Videos and nearly matches its video Consistency, suggesting that target semantics alone do not guarantee superior perceptual generation quality. We provide additional human evaluation details in Appendix~\ref{sec:human_eval}.

\begin{table}[t]
    \small
    \centering
    \caption{Ablation study on CID prediction modeling.}
    \label{tab:ablation-cid-modeling}
    \setlength{\tabcolsep}{2.3mm}
    \vspace{-0.12in}
    \begin{tabular}{c|cc|cc}
      \hline
      \multirow{2}{*}{Variant}
        & \multicolumn{2}{c|}{Games}
        & \multicolumn{2}{c}{Product} \\
      \cline{2-5}
        & R@20 & N@20 & R@20 & N@20 \\
      \hline
      w/o TID & 0.0546 & 0.0235 & 0.0199 & 0.0083 \\
    w/o Init & 0.0546 & 0.0242 & 0.0191 & 0.0079 \\
      w/o GRPO & 0.0613 & 0.0292 & 0.0198 & 0.0081 \\
      Init Only & 0.0530 & 0.0233 & 0.0092 & 0.0031 \\
       w/o Reasoning & 0.0604 & 0.0284 & 0.0167 & 0.0066 \\
       
      \hline
      \textbf{\model} & \textbf{0.0621} & \textbf{0.0294} & \textbf{0.0211} & \textbf{0.0088} \\
      \hline
    \end{tabular}
    \vspace{-0.18in}
\end{table}

\subsection{Ablation Study (RQ2)}

We conduct ablation studies from both personalized generation and CID-space collaborative modeling perspectives, with results shown in Figure~\ref{fig:ablation-image} and Table~\ref{tab:ablation-cid-modeling}. \textbf{Overall Effect.} The full \model\ variant achieves the strongest or most balanced results across generation relevance, novelty, and CID modeling metrics, indicating that the proposed modules contribute complementary signals. \textbf{Identifier Grounding.} Removing TID grounding or CID initialization weakens both generation and CID modeling performance, while the initialization-only variant performs worst, showing that stable collaborative identifier adaptation must be followed by semantic recovery and full collaborative trajectory modeling. \textbf{Reasoning and Reward Alignment.} Removing reasoning supervision hurts CID trajectory modeling and generation quality, and ablating GRPO leads to a weaker relevance--novelty trade-off and lower CID-space retrieval quality, confirming that transition-level reasoning and multi-task reward alignment jointly improve collaborative-sequence consistency and generation readiness.
\vspace{-0.05in}

\subsection{Collaborative Identifier Prediction (RQ3)}

We further evaluate \model\ on \textbf{CID2CID} prediction, where the model predicts the next CID from a user's historical CID sequence. This task is not intended to replace TID-based semantic prediction; rather, it isolates whether CID provides an additional view for discovering collaborative signals that are not directly exposed by textual identifiers. Table~\ref{tab:cid-modeling} reports Recall and NDCG on the three datasets. \model\ achieves the best results on most metrics, with clear advantages on Product and Short Videos, while OneRec remains stronger on the top-10 metrics of Games. Although \model\ is slightly weaker on Games top-10 retrieval, its advantage at R@20 and N@20 indicates broader top-k coverage in the collaborative identifier space, and its consistent gains on Product and Short Videos demonstrate robustness under different sparsity and item-distribution conditions.

\begin{figure}[t]
  \centering
  \begin{subfigure}[b]{0.48\linewidth}
    \centering
    \includegraphics[width=\linewidth]{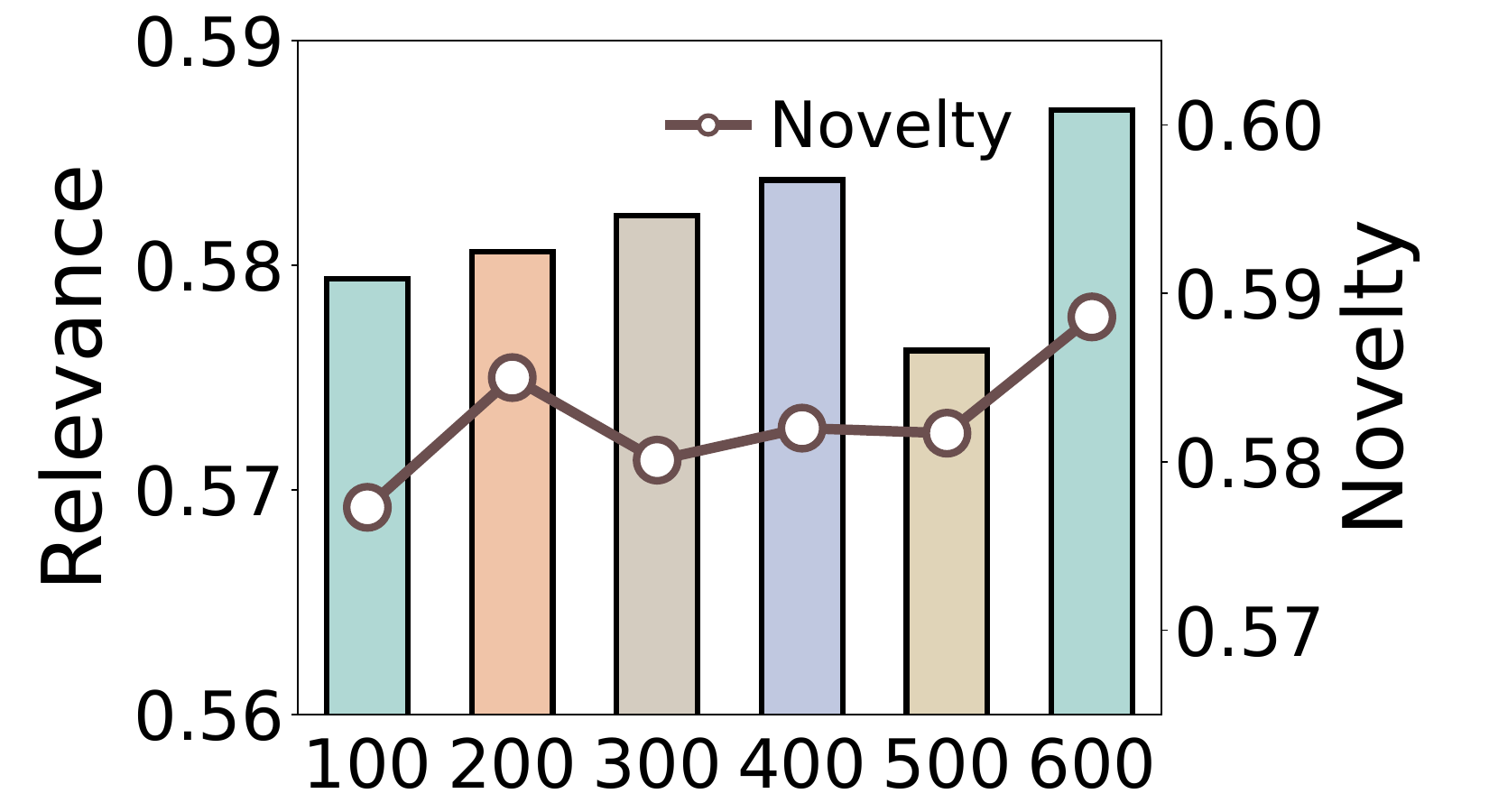}
    \vspace{-0.25in}
    \caption{Product Generation}
    \label{fig:hyperparam-product-generation}
  \end{subfigure}
  \hspace{0.01\linewidth}
  \begin{subfigure}[b]{0.48\linewidth}
    \centering
    \includegraphics[width=\linewidth]{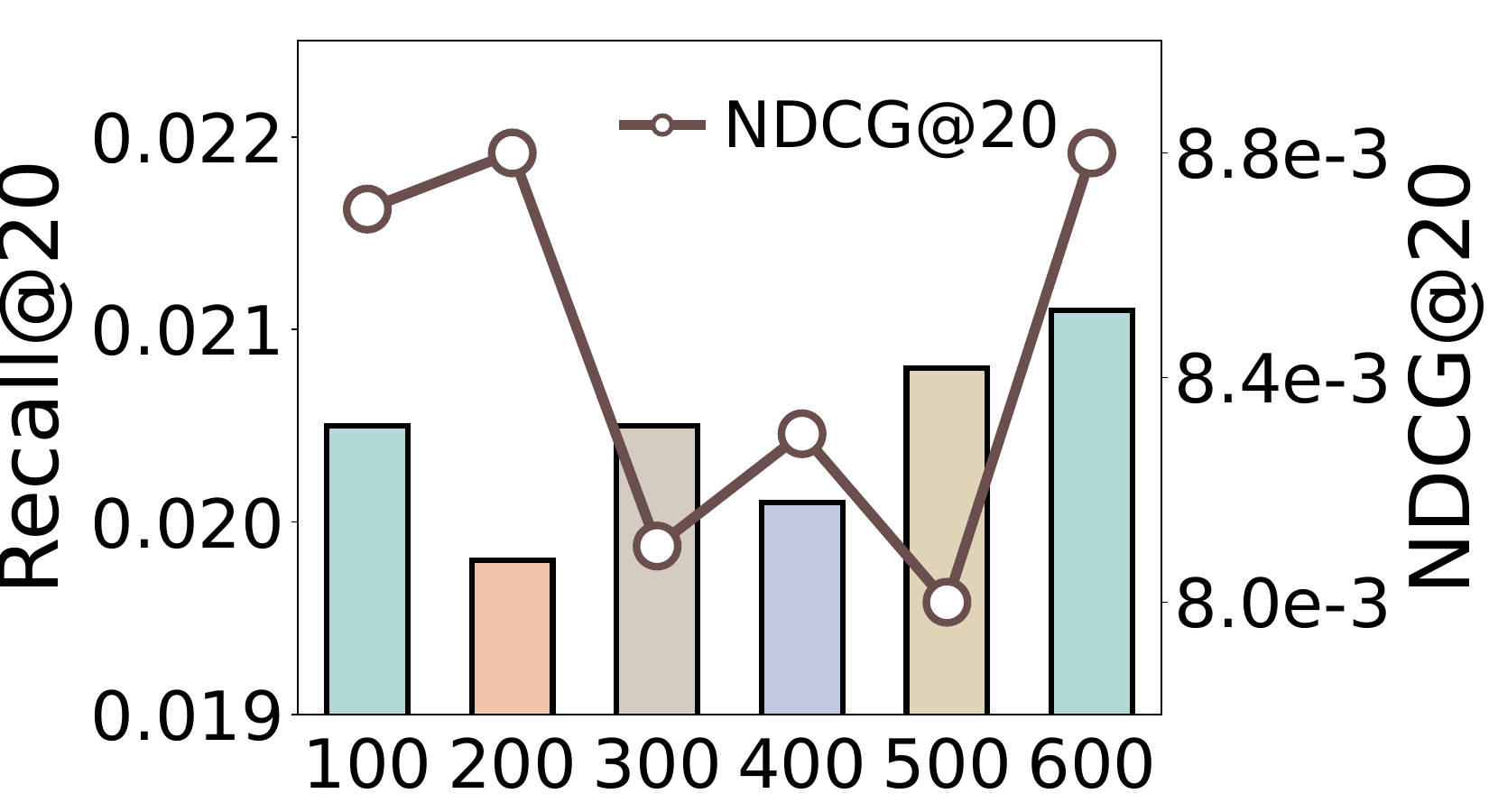}
    \vspace{-0.25in}
    \caption{Product Modeling}
    \label{fig:hyperparam-product-cid-modeling}
  \end{subfigure}
  \\[0.04in]
  \begin{subfigure}[b]{0.48\linewidth}
    \centering
    \includegraphics[width=\linewidth]{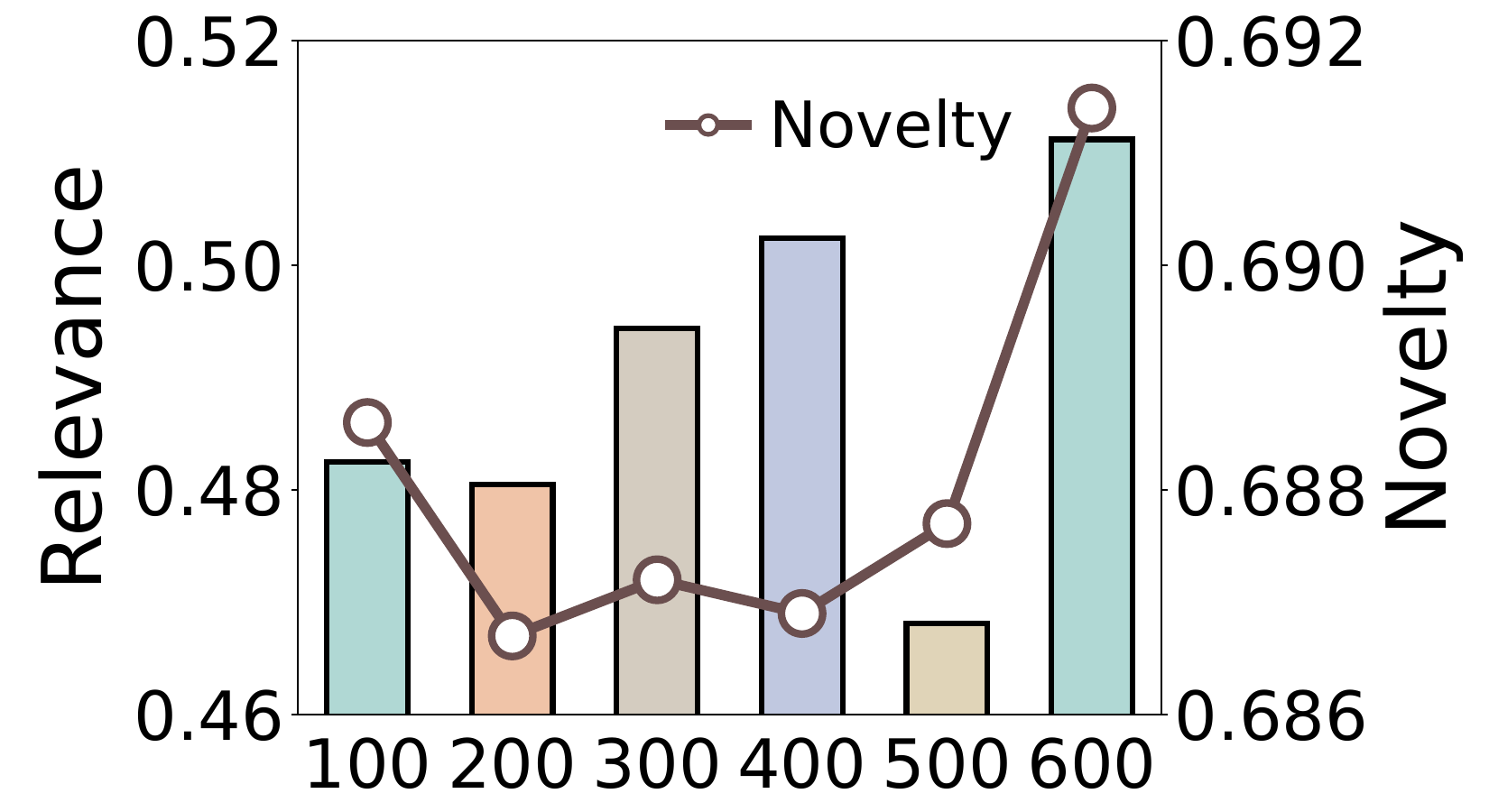}
    \vspace{-0.25in}
    \caption{Short Videos Generation}
    \label{fig:hyperparam-short-videos-generation}
  \end{subfigure}
  \hspace{0.01\linewidth}
  \begin{subfigure}[b]{0.48\linewidth}
    \centering
    \includegraphics[width=\linewidth]{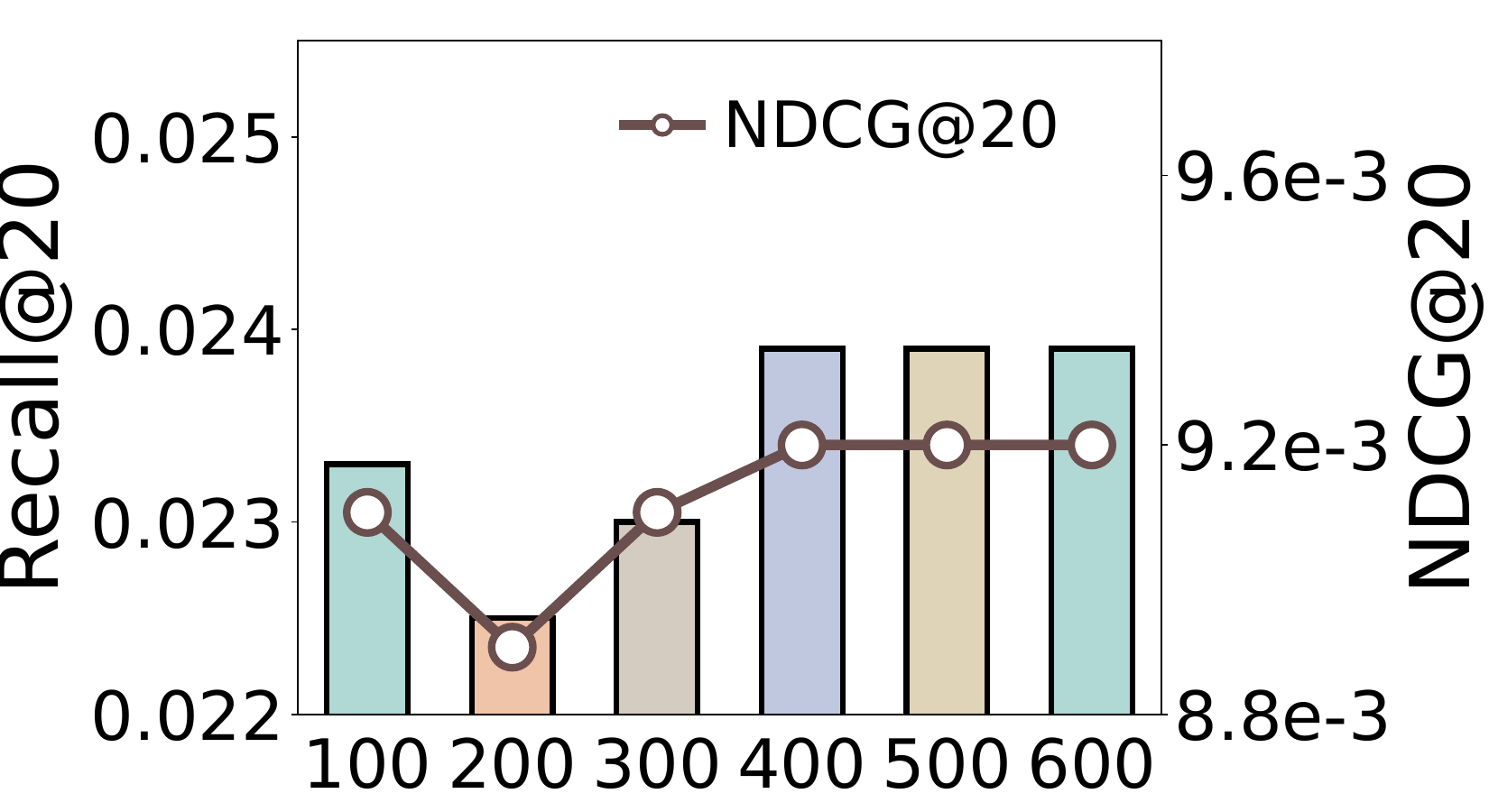}
    \vspace{-0.25in}
    \caption{Short Videos Modeling}
    \label{fig:hyperparam-short-videos-cid-modeling}
  \end{subfigure}
  \vspace{-0.28in}
  \caption{Hyperparameter study on GRPO steps.}
  \label{fig:hyperparameter-study}
  \vspace{-0.2in}
\end{figure}

\vspace{-0.05in}
\subsection{Hyperparameter Study (RQ4)}

We analyze the training-step sensitivity of \model\ in Table~\ref{tab:sft-hyperparameter} and Figure~\ref{fig:hyperparameter-study}. \textbf{SFT Steps.} Increasing supervised tuning mainly improves R@20 across datasets, while consistency only fluctuates within a narrow range, suggesting that longer SFT strengthens CID-space collaborative trajectory modeling without degrading instruction coherence. \textbf{GRPO Steps.} RL updates have a clearer effect on generation quality: relevance generally improves toward later steps and novelty remains stable or slightly increases, despite minor mid-training fluctuations. \textbf{CID Prediction Stability.} CID collaborative modeling metrics are less sensitive to GRPO, showing mild Product fluctuations and a Short Videos plateau, suggesting that later RL mainly affects generation-side alignment rather than substantially changing CID-space collaborative modeling.

\subsection{Case Study (RQ5)}
\begin{figure}[t] 
    \centering
    \includegraphics[width=\linewidth]{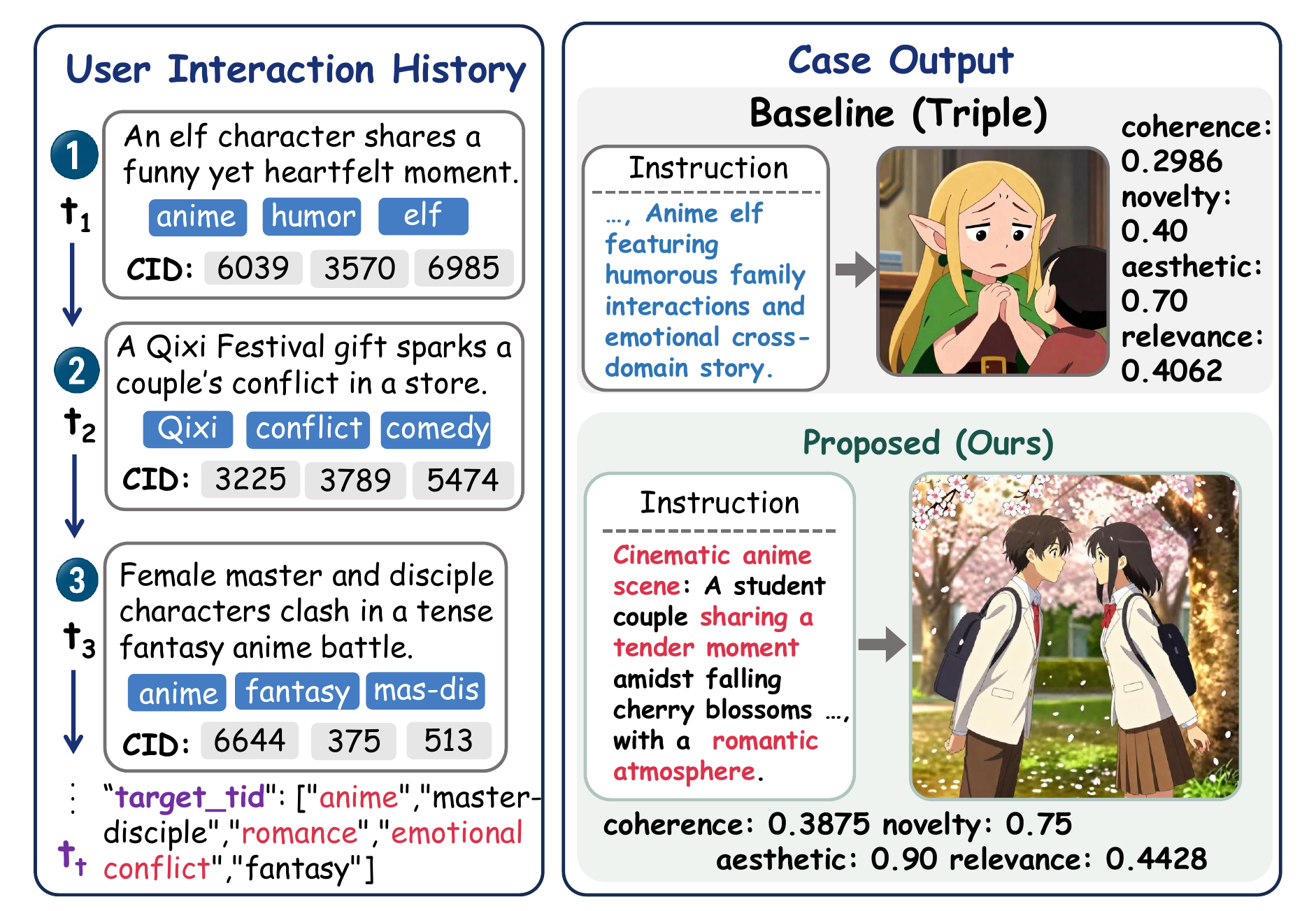}
    \caption{Case study on TRIPLE and our \model}
    \label{fig:casestudy}
    \vspace{-0.2in}
\end{figure}

Figure~\ref{fig:casestudy} presents a qualitative comparison with TRIPLE on history-guided generation. The user's history evolves from humorous elf interactions to conflict-centered fantasy anime, while the target TID further introduces romance and emotional conflict. TRIPLE captures coarse historical cues such as anime and elf, but its instruction remains generic and fails to reflect the target-side romantic transition. In contrast, \model\ leverages CID-level collaborative interaction transitions and TID-level semantic grounding to infer the next-interest direction, preserving the anime/fantasy context while specifying visual cues such as a student couple, a tender moment, and a romantic atmosphere. The image achieves higher scores across all metrics, suggesting \model\ distills collaborative interaction trajectories into generation-ready visual conditions rather than merely copying historical topics.

\begin{table}[t]
    \small
    \centering
    \caption{Hyperparameter study on SFT checkpoint steps. Ours corresponds to 7128 steps.}
    \label{tab:sft-hyperparameter}
    \setlength{\tabcolsep}{1.1mm}
    \renewcommand{\arraystretch}{1.15}
    \vspace{-0.12in}
    \begin{tabular}{c|cc|cc|cc}
      \hline
      \multirow{2}{*}{Steps}
        & \multicolumn{2}{c|}{Games}
        & \multicolumn{2}{c|}{Product}
        & \multicolumn{2}{c}{Short Videos} \\
      \cline{2-7}
        & Cons. & R@20 & Cons. & R@20 & Cons. & R@20 \\
      \hline
      2000 & 0.3499 & 0.0554 & 0.3243 & 0.0201 & \textbf{0.3450} & 0.0554 \\
      4000 & 0.3429 & 0.0613 & 0.3325 & 0.0198 & 0.3425 & 0.0613 \\
      6000 & 0.3502 & 0.0617 & \textbf{0.3328} & 0.0219 & \textbf{0.3450} & 0.0617 \\
      \hline
      Ours & \textbf{0.3505} & \textbf{0.0629} & 0.3319 & \textbf{0.0232} & 0.3409 & \textbf{0.0629} \\
      \hline
    \end{tabular}
\end{table}

\vspace{-0.05in}
\section{Related Work}
\label{sec:relate}

\noindent\textbf{Personalized Generation}.
Personalized generation adapts content to individual users, styles, or contexts, with one line personalizing visual synthesis from histories, retrieved evidence, or recommendation-guided signals ~\citep{shen2024pmg,xu2025personalized,ling2026ragar}. Another line constructs user-specific profiles or alignment signals from edits, demonstrations, or behavioral theories ~\citep{gao2024aligning,prose,noh2026triple}. While these methods improve user responsiveness, they often rely on profile-like descriptions rather than generation-ready instructions and provide limited coupling between compact behavioral modeling and explicit semantic grounding; in contrast, \model\ converts implicit interaction histories into executable AIGC instructions for downstream image and video synthesis.\\\vspace{-0.2in}

\noindent\textbf{Multimodal Generative Models}.
Multimodal generative models provide the backbone for turning natural-language instructions into images, videos, and other visual media ~\citep{xie2025show, huang2025interleaving}. VLMs increasingly support reasoning and content creation~\citep{hurst2024gpt, jin2025large}, while AIGC systems continue to improve prompt alignment, visual fidelity, temporal coherence, and motion control~\citep{sun2025content,seedream2025seedream,seedance2026seedance}. However, these models are mainly optimized to follow explicit prompts rather than infer generation conditions from sparse and noisy user behaviors; \model\ is therefore complementary, translating behavior-grounded preferences into executable instructions for downstream multimodal synthesis.\\\vspace{-0.12in}

\noindent\textbf{Behavioral Preference Modeling}.
Preference modeling learns user interests from behavioral data, with graph and intent-aware models capturing higher-order user--item relations~\citep{he2020lightgcn,zhang2024exploring,zhao2025symmetric}, and sequential or multimodal methods modeling temporal dynamics and heterogeneous content features~\citep{ren2024representation,fu2025efficient,lin2025contrastive}. Recent LLM-augmented methods further use language models to interpret item descriptions, refine candidate scoring, simulate users, or calibrate preference decisions~\citep{sun2023chatgpt,MoRE,ye2025harnessing}. \model\ instead transforms behavior-derived preference signals into personalized creation by connecting behavior-aware identifier prediction with generation-ready instruction.

\vspace{-0.05in}
\section{Conclusion}

We presented \model, a behavior-aware framework that turns user interaction history into executable instructions for personalized AIGC. Built on a dual-identifier item representation, \model\ couples reasoning-augmented supervised tuning with multi-task reinforcement learning to bridge user preference modeling and downstream generation control. By separating CID-based behavioral coding from TID-based semantic grounding, \model\ treats preference prediction as an intermediate reasoning step rather than an end in itself. Experiments across product, game, and short-video domains demonstrate consistent improvements in personalized image and video generation, instruction quality, and collaborative identifier prediction,  establishing a unified path from implicit user behavior to controllable multimodal synthesis.


\section{Limitations}

Personalized generation is bounded by how user preference evidence is collected and used in real applications. Interaction histories can contain sensitive behavioral signals, and users may not always expect such signals to be transformed into creative generation conditions. Practical deployments should therefore require explicit opt-in consent, minimize retained histories, anonymize or aggregate logs whenever possible, and provide clear controls for inspecting, editing, or deleting personalized preference profiles. We view personalized instruction generation as an assistive layer for creative control rather than a replacement for deployment-time safety governance.

\section{Ethical Considerations}

This work uses existing datasets and item captions for offline research evaluation, without collecting new personal data, inferring sensitive demographic attributes, or deploying personalized generation to real users. The generated instructions are evaluated only as experimental outputs for studying preference-to-instruction modeling. Beyond the privacy and deployment precautions discussed in the limitations, we do not identify additional ethical concerns specific to this work.

\bibliography{ref}

@article{onerec,
  title={{OneRec}: Unifying Retrieve and Rank with Generative Recommender and Iterative Preference Alignment},
  author={Deng, Jiaxin and Wang, Shiyao and Cai, Kuo and Ren, Lejian and Hu, Qigen and Ding, Weifeng and Luo, Qiang and Zhou, Guorui},
  journal={arXiv preprint arXiv:2502.18965},
  year={2025}
}

@inproceedings{lcrec,
  title={Adapting Large Language Models by Integrating Collaborative Semantics for Recommendation},
  author={Zheng, Bowen and Hou, Yupeng and Lu, Hongyu and Chen, Yu and Zhao, Wayne Xin and Chen, Ming and Wen, Ji-Rong},
  booktitle={2024 IEEE 40th International Conference on Data Engineering},
  pages={1435--1448},
  year={2024},
  publisher={IEEE},
  doi={10.1109/ICDE60146.2024.00118}
}

@inproceedings{he2020lightgcn,
  title={Lightgcn: Simplifying and powering graph convolution network for recommendation},
  author={He, Xiangnan and Deng, Kuan and Wang, Xiang and Li, Yan and Zhang, Yongdong and Wang, Meng},
  booktitle={SIGIR},
  pages={639--648},
  year={2020}
}

@inproceedings{zhang2024exploring,
  title={Exploring the individuality and collectivity of intents behind interactions for graph collaborative filtering},
  author={Zhang, Yi and Sang, Lei and Zhang, Yiwen},
  booktitle={Proceedings of the 47th International ACM SIGIR Conference on Research and Development in Information Retrieval},
  pages={1253--1262},
  year={2024}
}

@article{zhao2025symmetric,
  title={Symmetric Graph Contrastive Learning against Noisy Views for Recommendation},
  author={Zhao, Chu and Yang, Enneng and Liang, Yuliang and Zhao, Jianzhe and Guo, Guibing and Wang, Xingwei},
  journal={ACM TOIS},
  volume={43},
  number={3},
  pages={1--28},
  year={2025},
  publisher={ACM New York, NY}
}

@inproceedings{ren2024representation,
  title={Representation learning with large language models for recommendation},
  author={Ren, Xubin and Wei, Wei and Xia, Lianghao and Su, Lixin and Cheng, Suqi and Wang, Junfeng and Yin, Dawei and Huang, Chao},
  booktitle={WWW},
  pages={3464--3475},
  year={2024}
}

@article{fu2025efficient,
  title={Efficient and effective adaptation of multimodal foundation models in sequential recommendation},
  author={Fu, Junchen and Ge, Xuri and Xin, Xin and Karatzoglou, Alexandros and Arapakis, Ioannis and Zheng, Kaiwen and Ni, Yongxin and Joemon, Joemon M Jose},
  journal={IEEE TKDE},
  year={2025},
  publisher={IEEE}
}

@article{lin2025contrastive,
  title={Contrastive modality-disentangled learning for multimodal recommendation},
  author={Lin, Xixun and Liu, Rui and Cao, Yanan and Zou, Lixin and Li, Qian and Wu, Yongxuan and Liu, Yang and Yin, Dawei and Xu, Guandong},
  journal={ACM TOIS},
  volume={43},
  number={3},
  pages={1--31},
  year={2025},
  publisher={ACM New York, NY}
}

@article{sun2023chatgpt,
  title={Is ChatGPT good at search? investigating large language models as re-ranking agents},
  author={Sun, Weiwei and Yan, Lingyong and Ma, Xinyu and Wang, Shuaiqiang and Ren, Pengjie and Chen, Zhumin and Yin, Dawei and Ren, Zhaochun},
  journal={arXiv preprint arXiv:2304.09542},
  year={2023}
}

@inproceedings{MoRE,
  title={More: A mixture of reflectors framework for large language model-based sequential recommendation},
  author={Qin, Weicong and Xu, Yi and Yu, Weijie and Shen, Chenglei and Zhang, Xiao and He, Ming and Fan, Jianping and Xu, Jun},
  booktitle={Recsys},
  pages={299--308},
  year={2025}
}

@inproceedings{ye2025harnessing,
  title={Harnessing multimodal large language models for multimodal sequential recommendation},
  author={Ye, Yuyang and Zheng, Zhi and Shen, Yishan and Wang, Tianshu and Zhang, Hengruo and Zhu, Peijun and Yu, Runlong and Zhang, Kai and Xiong, Hui},
  booktitle={AAAI},
  volume={39},
  number={12},
  pages={13069--13077},
  year={2025}
}

@inproceedings{prose,
  title={Aligning {LLM}s by Predicting Preferences from User Writing Samples},
  author={Aroca-Ouellette, St{\'e}phane and Mackraz, Natalie and Theobald, Barry-John and Metcalf, Katherine},
  booktitle={Proceedings of the 42nd International Conference on Machine Learning},
  pages={1690--1721},
  year={2025},
  volume={267},
  series={Proceedings of Machine Learning Research},
  publisher={PMLR}
}

@article{dang2026multi,
  title={Multi-agent collaboration via evolving orchestration},
  author={Dang, Yufan and Qian, Chen and Luo, Xueheng and Fan, Jingru and Xie, Zihao and Shi, Ruijie and Chen, Weize and Yang, Cheng and Che, Xiaoyin and Tian, Ye and others},
  journal={Advances in neural information processing systems},
  volume={38},
  pages={165025--165059},
  year={2026}
}

@article{hurst2024gpt,
  title={Gpt-4o system card},
  author={Hurst, Aaron and Lerer, Adam and Goucher, Adam P and Perelman, Adam and Ramesh, Aditya and Clark, Aidan and Ostrow, AJ and Welihinda, Akila and Hayes, Alan and Radford, Alec and others},
  journal={arXiv preprint arXiv:2410.21276},
  year={2024}
}

@inproceedings{jin2025large,
  title={Large Vison-Language Foundation Model in Baidu AIGC Image Advertising},
  author={Jin, Zhipeng and Tao, Wen and Li, Yafei and Yang, Yi and Han, Cong and Li, Shuanglong and Liu, Lin},
  booktitle={KDD},
  pages={2303--2312},
  year={2025}
}

@article{sun2025content,
  title={Content-rich aigc video quality assessment via intricate text alignment and motion-aware consistency},
  author={Sun, Shangkun and Liang, Xiaoyu and Qu, Bowen and Gao, Wei},
  journal={arXiv preprint arXiv:2502.04076},
  year={2025}
}

@article{seedream2025seedream,
  title={Seedream 4.0: Toward next-generation multimodal image generation},
  author={Seedream, Team and Chen, Yunpeng and Gao, Yu and Gong, Lixue and Guo, Meng and Guo, Qiushan and Guo, Zhiyao and Hou, Xiaoxia and Huang, Weilin and Huang, Yixuan and others},
  journal={arXiv preprint arXiv:2509.20427},
  year={2025}
}

@inproceedings{yang2025cogvideox,
  title={Cogvideox: Text-to-video diffusion models with an expert transformer},
  author={Yang, Zhuoyi and Teng, Jiayan and Zheng, Wendi and Ding, Ming and Huang, Shiyu and Xu, Jiazheng and Yang, Yuanming and Hong, Wenyi and Zhang, Xiaohan and Feng, Guanyu and others},
  booktitle={International Conference on Learning Representations},
  volume={2025},
  pages={83048--83077},
  year={2025}
}

@article{seedance2026seedance,
  title={Seedance 2.0: Advancing video generation for world complexity},
  author={Seedance, Team and Chen, De and Chen, Liyang and Chen, Xin and Chen, Ying and Chen, Zhuo and Chen, Zhuowei and Cheng, Feng and Cheng, Tianheng and Cheng, Yufeng and others},
  journal={arXiv preprint arXiv:2604.14148},
  year={2026}
}

@article{huang2025interleaving,
  title={Interleaving reasoning for better text-to-image generation},
  author={Huang, Wenxuan and Chen, Shuang and Xie, Zheyong and Cao, Shaosheng and Tang, Shixiang and Shen, Yufan and Yin, Qingyu and Hu, Wenbo and Wang, Xiaoman and Tang, Yuntian and others},
  journal={arXiv preprint arXiv:2509.06945},
  year={2025}
}

@inproceedings{xie2025show,
  title={Show-o: One single transformer to unify multimodal understanding and generation},
  author={Xie, Jinheng and Mao, Weijia and Bai, Zechen and Zhang, David Junhao and Wang, Weihao and Lin, Kevin Qinghong and Gu, Yuchao and Chen, Zhijie and Yang, Zhenheng and Shou, Mike Zheng},
  booktitle={International Conference on Learning Representations},
  volume={2025},
  pages={28240--28264},
  year={2025}
}

@inproceedings{zhao2025local,
  title={Local conditional controlling for text-to-image diffusion models},
  author={Zhao, Yibo and Peng, Liang and Yang, Yang and Luo, Zekai and Li, Hengjia and Chen, Yao and Yang, Zheng and He, Xiaofei and Zhao, Wei and Lu, Qinglin and others},
  booktitle={Proceedings of the AAAI conference on artificial intelligence},
  volume={39},
  number={10},
  pages={10492--10500},
  year={2025}
}

@inproceedings{ling2026ragar,
  title={RAGAR: retrieval augmented personalized image generation guided by recommendation},
  author={Ling, Run and Wang, Wenji and Liu, Yuting and Guo, Guibing and Liu, Haowei and Lu, Jian and Zhang, Quanwei and Xu, Yexing and Lu, Shuo and Wang, Yun},
  booktitle={Proceedings of the AAAI Conference on Artificial Intelligence},
  volume={40},
  number={18},
  pages={15278--15286},
  year={2026}
}

@article{an2026unictokens,
  title={Unictokens: Boosting personalized understanding and generation via unified concept tokens},
  author={An, Ruichuan and Yang, Sihan and Zhang, Renrui and Lu, Ming and Dai, Gaole and Liang, Hao and Guo, Ziyu and Yan, Shilin and Luo, Yulin and Zou, Bocheng},
  journal={Advances in Neural Information Processing Systems},
  volume={38},
  pages={144638--144664},
  year={2026}
}

@inproceedings{xu2025personalized,
  title={Personalized image generation with large multimodal models},
  author={Xu, Yiyan and Wang, Wenjie and Zhang, Yang and Tang, Biao and Yan, Peng and Feng, Fuli and He, Xiangnan},
  booktitle={Proceedings of the ACM on Web Conference 2025},
  pages={264--274},
  year={2025}
}

@article{zhang2026unleashing,
  title={Unleashing the Native Recommendation Potential: LLM-Based Generative Recommendation via Structured Term Identifiers},
  author={Zhang, Zhiyang and She, Junda and Cai, Kuo and Chen, Bo and Wang, Shiyao and Luo, Xinchen and Luo, Qiang and Tang, Ruiming and Li, Han and Gai, Kun and others},
  journal={arXiv preprint arXiv:2601.06798},
  year={2026}
}

@article{kao2017deep,
  title={Deep aesthetic quality assessment with semantic information},
  author={Kao, Yueying and He, Ran and Huang, Kaiqi},
  journal={TIP},
  volume={26},
  number={3},
  pages={1482--1495},
  year={2017},
  publisher={IEEE}
}

@inproceedings{vargas2011rank,
  title={Rank and relevance in novelty and diversity metrics for recommender systems},
  author={Vargas, Sa{\'u}l and Castells, Pablo},
  booktitle={Recsys},
  pages={109--116},
  year={2011}
}

@inproceedings{hessel2021clipscore,
  title={Clipscore: A reference-free evaluation metric for image captioning},
  author={Hessel, Jack and Holtzman, Ari and Forbes, Maxwell and Le Bras, Ronan and Choi, Yejin},
  booktitle={EMNLP},
  pages={7514--7528},
  year={2021}
}

@article{gao2024aligning,
  title={Aligning llm agents by learning latent preference from user edits},
  author={Gao, Ge and Taymanov, Alexey and Salinas, Eduardo and Mineiro, Paul and Misra, Dipendra},
  journal={Advances in neural information processing systems},
  volume={37},
  pages={136873--136896},
  year={2024}
}

@inproceedings{noh2026triple,
  author    = {Noh, Taehyung and Jin, Seungwan and Yeo, Haein and Han, Kyungsik},
  title     = {TRIPLE: Theory-Driven Integration of Planned and Habitual Behaviors for LLM-based Personalization},
  booktitle = {Proceedings of the 40th AAAI Conference on Artificial Intelligence (AAAI-26)},
  year      = {2026},
  publisher = {AAAI Press}
}

@inproceedings{shen2024pmg,
  title={Pmg: Personalized multimodal generation with large language models},
  author={Shen, Xiaoteng and Zhang, Rui and Zhao, Xiaoyan and Zhu, Jieming and Xiao, Xi},
  booktitle={Proceedings of the ACM Web Conference 2024},
  pages={3833--3843},
  year={2024}
}

@inproceedings{kang2018self,
  title={Self-attentive sequential recommendation},
  author={Kang, Wang-Cheng and McAuley, Julian},
  booktitle={2018 IEEE international conference on data mining (ICDM)},
  pages={197--206},
  year={2018},
  organization={IEEE}
}

@article{zhou2025openonerec,
  title={OpenOneRec Technical Report},
  author={Zhou, Guorui and Bao, Honghui and Huang, Jiaming and Deng, Jiaxin and Zhang, Jinghao and She, Junda and Cai, Kuo and Ren, Lejian and Ren, Lu and Luo, Qiang and others},
  journal={arXiv preprint arXiv:2512.24762},
  year={2025}
}

@article{rajput2023recommender,
  title={Recommender systems with generative retrieval},
  author={Rajput, Shashank and Mehta, Nikhil and Singh, Anima and Hulikal Keshavan, Raghunandan and Vu, Trung and Heldt, Lukasz and Hong, Lichan and Tay, Yi and Tran, Vinh and Samost, Jonah and others},
  journal={Advances in Neural Information Processing Systems},
  volume={36},
  pages={10299--10315},
  year={2023}
}

\clearpage
\appendix
\begin{figure*}[t]
    \centering
    \captionsetup[subfigure]{skip=0pt}

    \begin{subfigure}{\textwidth}
        \centering
        \includegraphics[width=\textwidth]{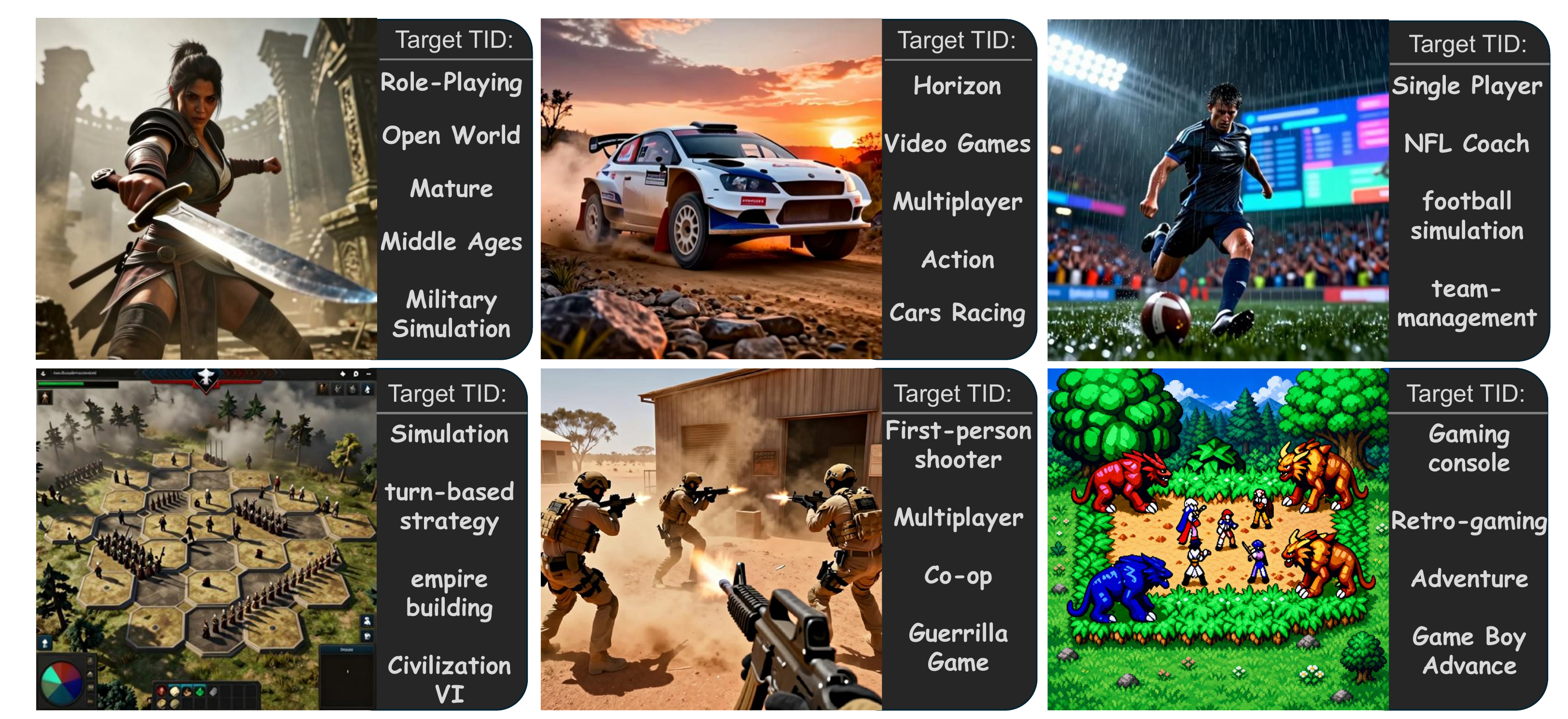}
        \vspace{-0.15in}
        \caption{Image: Game}
        \label{fig:case_game}
    \end{subfigure}
    
    \vspace{-1mm}

    \begin{subfigure}{\textwidth}
        \centering
        \includegraphics[width=\textwidth]{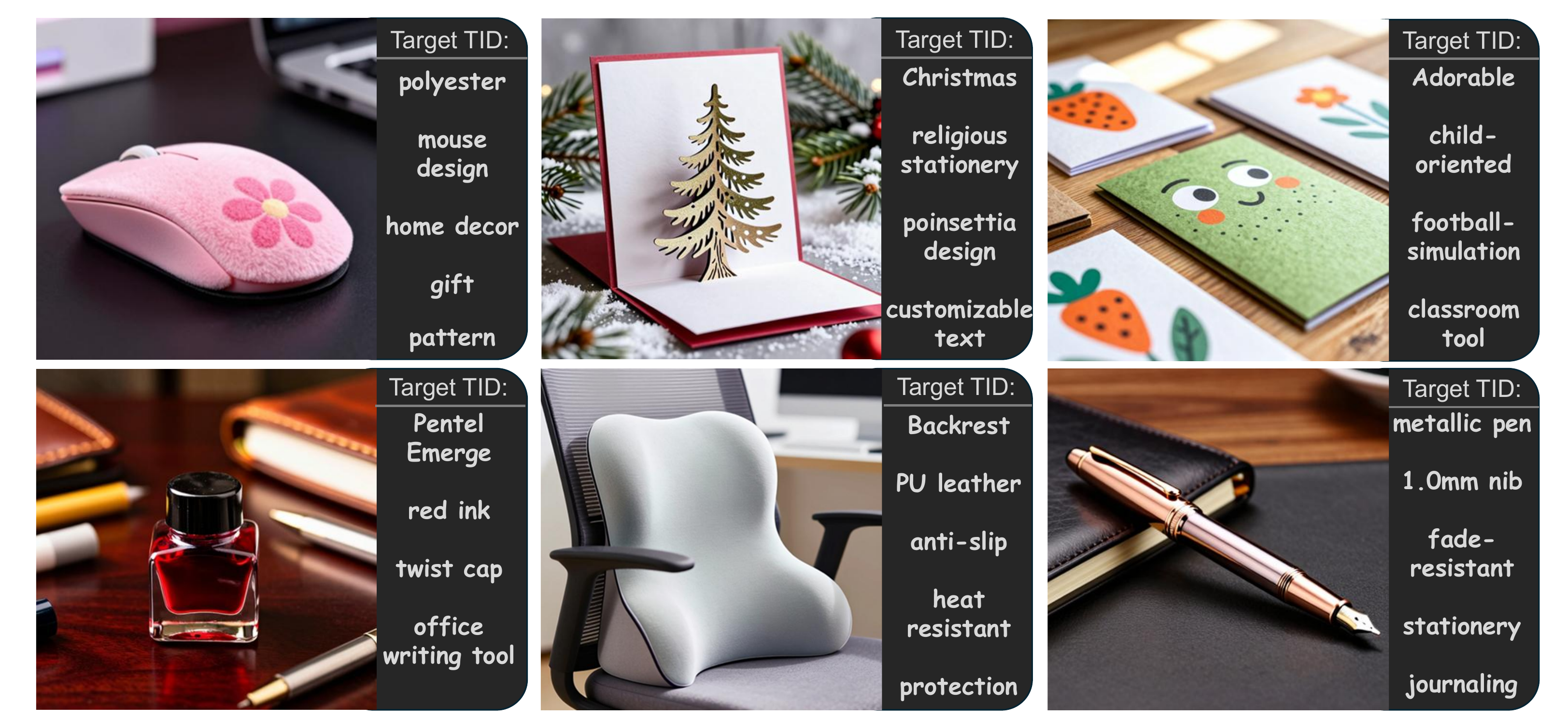}
        \vspace{-0.15in}
        \caption{Image: Product}
        \label{fig:case_product}
    \end{subfigure}

    \begin{subfigure}{\textwidth}
        \centering
        \includegraphics[width=\textwidth]{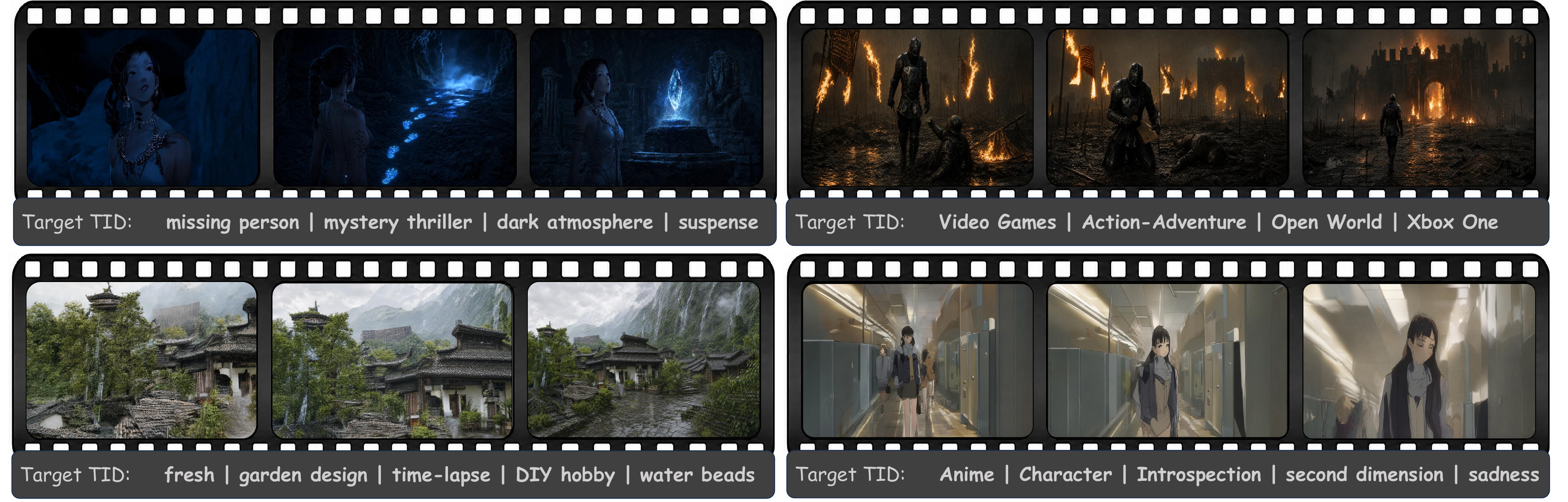}
        \vspace{-0.1in}
        \caption{Video: Short Video}
        \label{fig:case_video}
    \end{subfigure}

    \vspace{-2mm}
    \caption{Comprehensive generation cases across three representative domains, each with its target tid.}
    \label{fig:overall_case_studies}
\end{figure*}

\section{Appendix}

\subsection{Baseline Methods}
  To ensure a comprehensive study, we compare \model\ against a broad set of baselines covering personalized generation, collaborative filtering and broader behavioral preference modeling, and prompting-based references.

  \textbf{Personalized Generation Methods}
  \setlength{\leftmargini}{10pt}
  \begin{itemize}
   \item {\textbf{PMG}} \cite{shen2024pmg}: LLM-extracted user cues condition multimodal generators for personalized content synthesis, guided by behavior-aware prompts.
  \item {\textbf{PIGEON}} \cite{xu2025personalized}: Retrieved preference evidence steers frozen generation agents without model fine-tuning, preserving efficient deployment workflows.
  \item {\textbf{RAGAR}} \cite{ling2026ragar}: Semantic retrieval weights relevant histories, while ranking feedback balances personalization and fidelity across interaction-rich scenarios.
  \ \item {\textbf{CIPHER}} 
~\citep{gao2024aligning}: Historical user edits are retrieved to infer preferences and align generated outputs with user intent through edit-aware context matching.
  \item {\textbf{PROSE}} 
  ~\citep{prose}: Iterative refinement and consistency checks infer user preferences from demonstrations, using structured self-verification loops.
  \item {\textbf{TRIPLE}}
  ~\citep{noh2026triple}: A theory-driven LLM-based personalization framework integrating planned and habitual behavior modeling for preference reasoning.
  \end{itemize}

  \textbf{Behavioral Preference Modeling Methods}
  \setlength{\leftmargini}{10pt}
  \begin{itemize} 
  \item {\textbf{SASRec}}
  ~\citep{kang2018self}: Transformer-based sequential preference model learning user behavior from item ID sequences via self-attention.
  \item {\textbf{TIGER}}
  \cite{rajput2023recommender}: A generative item-ID modeling framework that converts item IDs into tokens and models sequential preference prediction as next-token generation.
  \item {\textbf{LC-Rec}}
  ~\citep{lcrec}: An ID-based sequential preference modeling method that enhances representation learning with additional latent/contextual signals.
  \begin{figure}[H]
  \centering
  
  \includegraphics[width=0.6\linewidth]{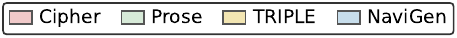}

  \begin{minipage}[b]{\linewidth}
    \centering
    \includegraphics[width=\linewidth]{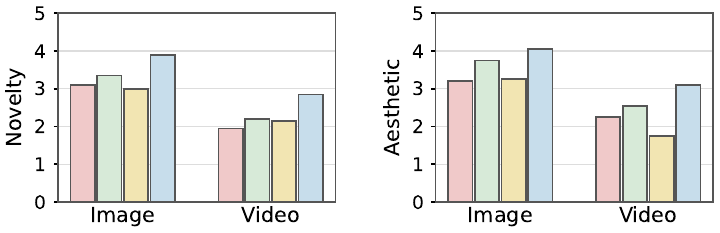}
  \end{minipage}
  \vspace{-0.1in}
  \caption{Human evaluation on created content.}
  \label{fig:human_eval}
  \vspace{-0.1in}
\end{figure}
  \item {\textbf{OpenOneRec}}
    ~\citep{zhou2025openonerec}: It integrates item-text alignment into an end-to-end generative preference modeling framework for scalable preference prediction and reasoning.  
  \end{itemize}

  \textbf{Prompting and Reference Baselines}
  \setlength{\leftmargini}{10pt}
  \begin{itemize}
   \item {\textbf{NPC}}
  : No-preference conditioning removes user evidence, such as reference images and similar historical items, using a generic prompt for generation to isolate personalization effects.
  \item {\textbf{Oracle}}
  : Ground-truth target semantics, such as the target caption or TID, are used as a non-deployable upper-bound reference for diagnostic comparison only.
  \end{itemize}

\subsection{Human Evaluation on Content}
\label{sec:human_eval}

To complement automatic VLM-based evaluation, we further conduct a human study to directly assess the perceptual quality of NaviGen. We perform human evaluation which averaged over three domains, including Product, Short Videos, and Games. From each dataset, we randomly sample 20 image cases and 5 video cases for evaluation. For NaviGen and three representative generation baselines, CIPHER, PROSE, and TRIPLE, which all support both image and video generation, we recruit 24 student volunteers to evaluate anonymized and randomly shuffled outputs using a 5-point Likert scale along two dimensions: Novelty, reflecting the creativity and non-triviality of the visual interpretation, and Aesthetic quality, capturing visual appeal, composition, clarity, and overall polish. Scores from 1 to 5 indicate very poor, poor, fair, good, and excellent quality, respectively. The evaluation process takes approximately 1.5 hours to complete. As shown in Figure~\ref{fig:human_eval}, NaviGen consistently achieves higher average ratings on both dimensions in image and video settings, suggesting that behavior-conditioned instructions lead to more creative and visually appealing personalized content.

\clearpage
\subsection{Prompt Templates}
\label{sec:prompt_temp}
\begin{figure}[h] 
    \centering
    \includegraphics[width=\linewidth]{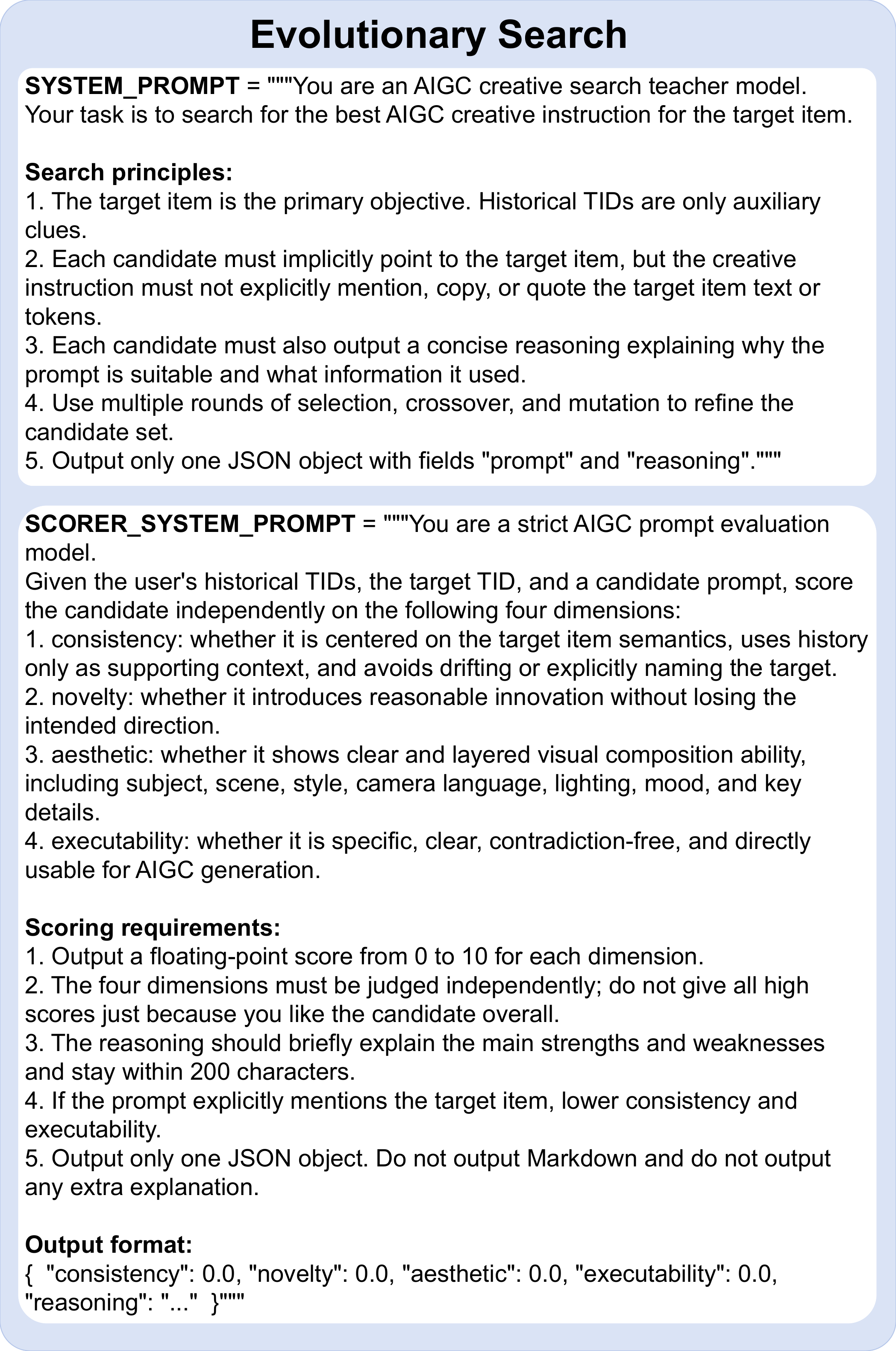}
    \caption{Generation prompts create target-aligned AIGC candidates, while scoring prompts evaluate and select the best final prompt.}
    \label{fig:evol}
\end{figure}

\begin{figure}[t]
    \centering
    \includegraphics[width=\linewidth]{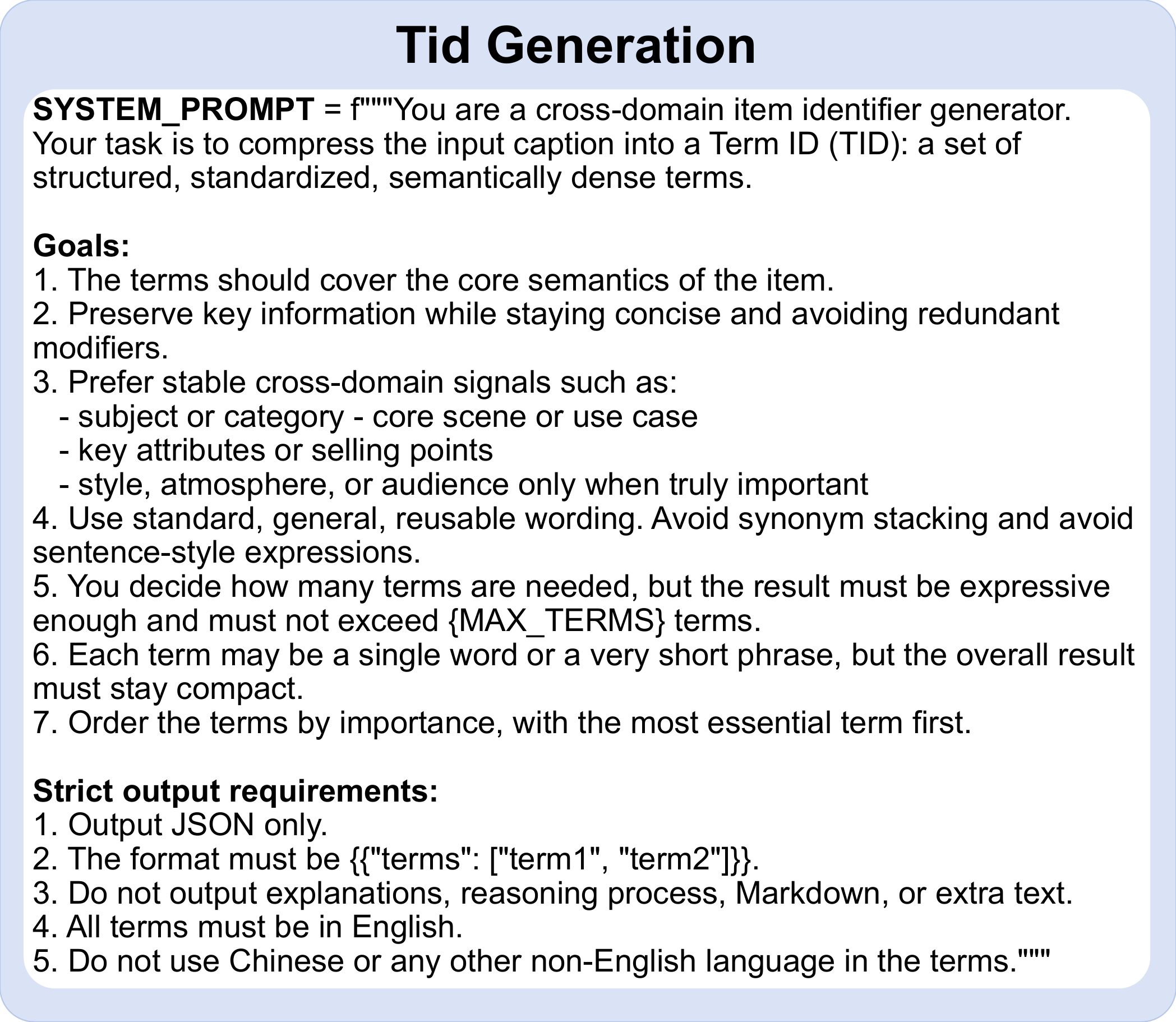}
    \caption{Convert item captions into structured Term IDs.}
    \label{fig:tid}
\end{figure}

\begin{figure}[t] 
    \centering
    \includegraphics[width=\linewidth]{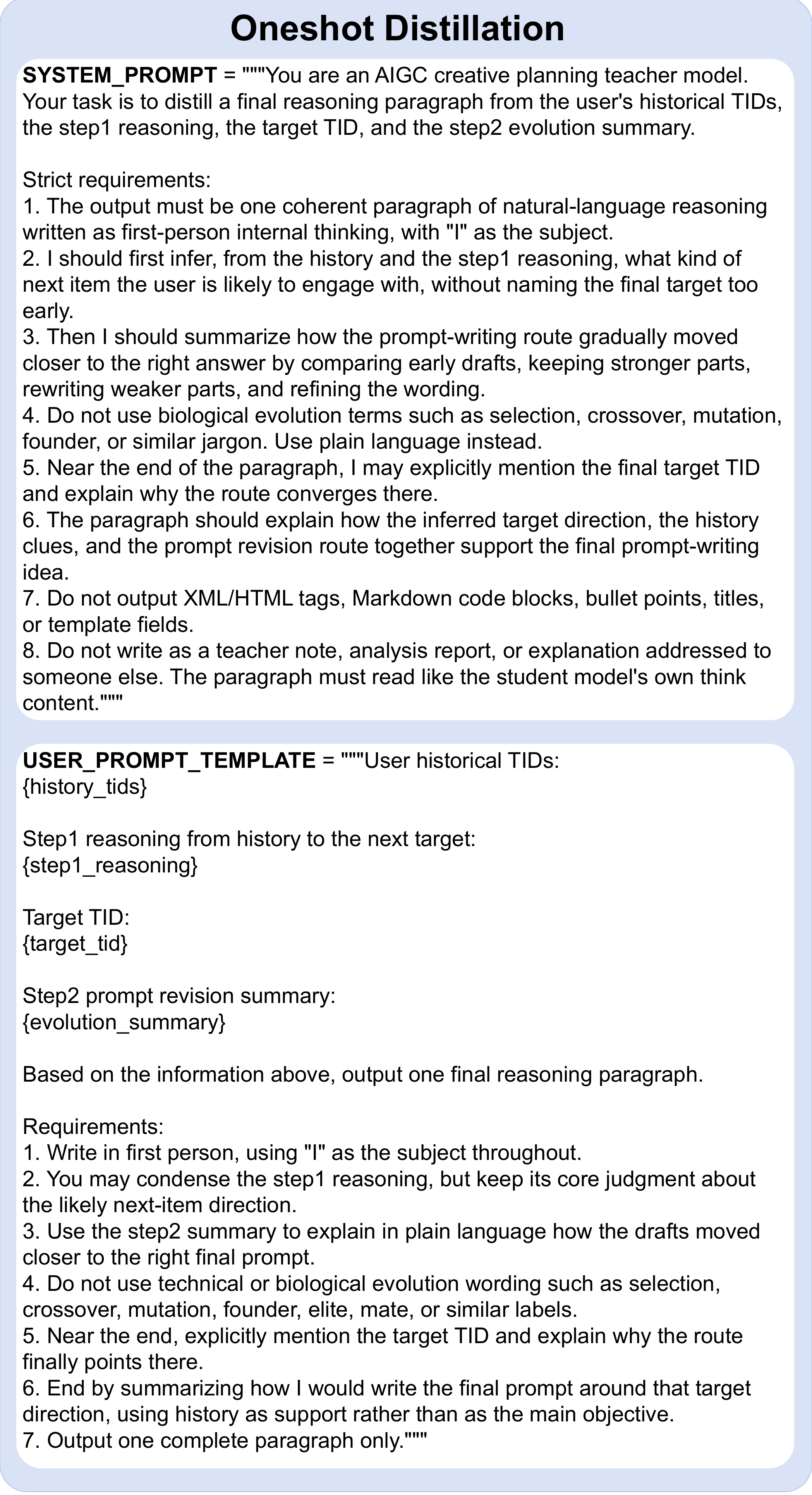}
    \caption{Distills user history, target reasoning, and prompt refinement into final first-person reasoning.}
    \label{fig:dist}
\end{figure}

\begin{figure*}[t] 
    \centering
    \vspace{-10mm}
    \includegraphics[width=\linewidth]{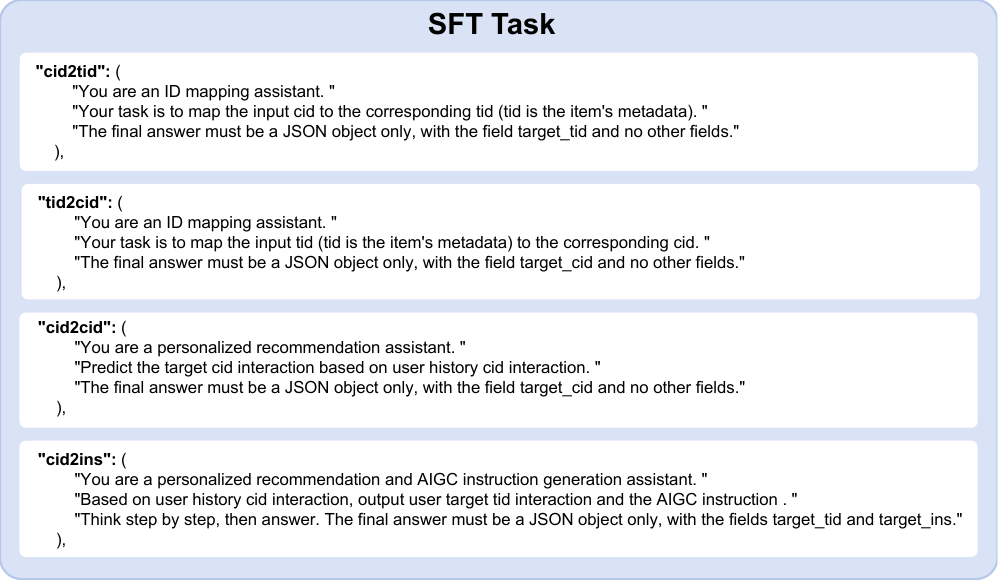}
    \caption{Four prompts of SFT tasks' construction.}
    \label{fig:sft}
\end{figure*}

\begin{figure*}[t] 
    \centering
    \vspace{-5mm}
    \includegraphics[width=\linewidth]{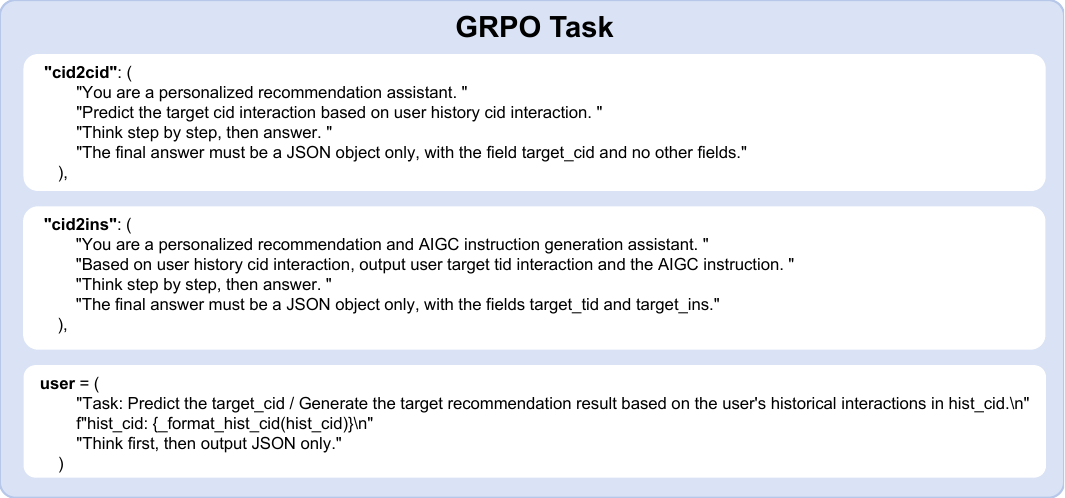}
    \caption{Two prompts of GRPO tasks' construction.}
    \label{fig:grpo}
\end{figure*}

\begin{figure*}[t] 
    \centering
    \includegraphics[width=\linewidth]{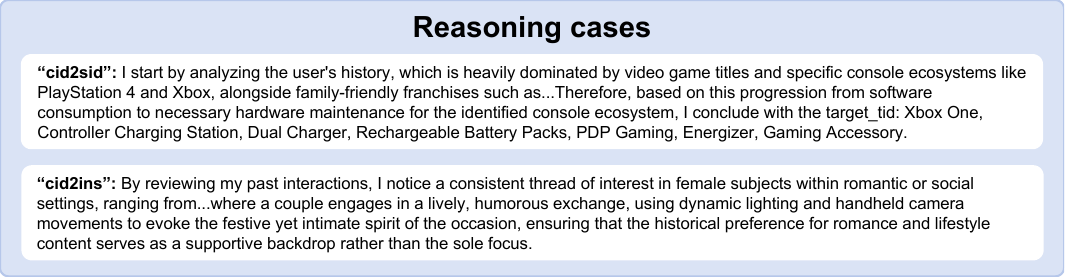}
    \caption{Cases of reasoning}
    \label{fig:reason}
\end{figure*}

\begin{figure*}[t] 
    \centering
    \includegraphics[width=\linewidth]{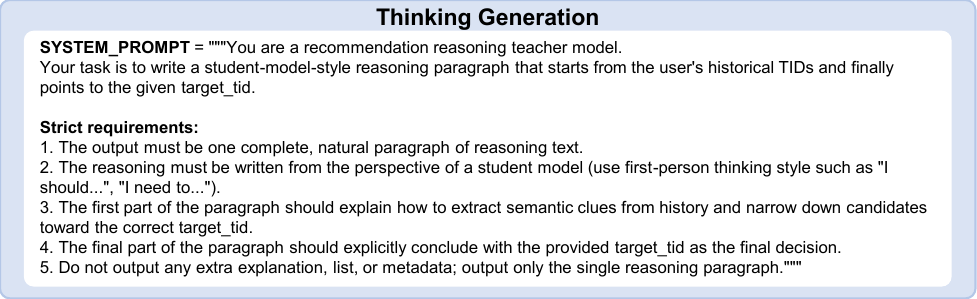}
    \caption{Reason from hist TIDs to target TID.}
    \label{fig:think}
\end{figure*}

\begin{figure*}[t]
    \centering
    \begin{subfigure}[b]{\linewidth}
        \centering
        \includegraphics[width=\linewidth]{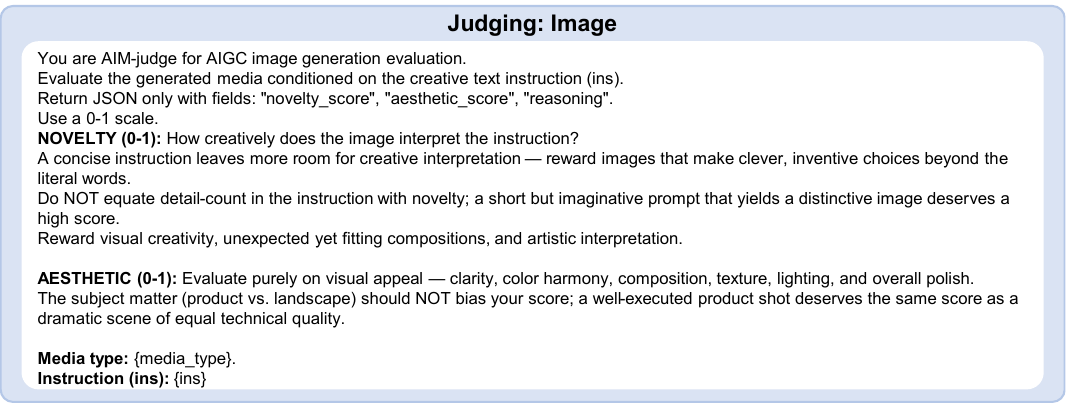}
        \caption{Image generation evaluation}
        \label{fig:judge_image}
    \end{subfigure}
    \\[2ex] 
    \begin{subfigure}[b]{\linewidth}
        \centering
        \includegraphics[width=\linewidth]{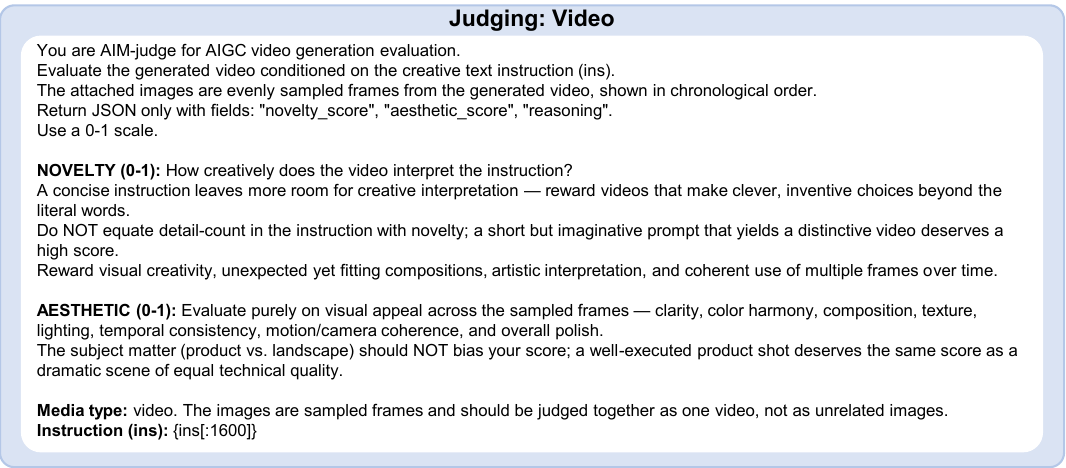}
        \caption{Video generation evaluation}
        \label{fig:judge_video}
    \end{subfigure}
    
    \caption{AIM-judge prompts for novelty and aesthetic evaluation. Both static images (a) and dynamic video sequences (b) are appraised for their creative interpretation of instructions (novelty) and visual polish (aesthetics).}
    \label{fig:overall_judging}
\end{figure*}

We apply these prompts to six main works and two reasoning cases (shown at last), including:
\begin{itemize}[noitemsep, leftmargin=*]
    \item \textbf{TID Generation:} Fig.~\ref{fig:tid} converts item captions into compact semantic TIDs.
    \item \textbf{Thinking Generation:} Fig.~\ref{fig:think} generates reasoning from historical TIDs to the target TID.
    \item \textbf{Evolutionary Search:} Fig.~\ref{fig:evol} searches for the best target-aligned AIGC prompt.
    \item \textbf{Oneshot Distillation:} Fig.~\ref{fig:dist} distills history and so on into final first-person reasoning.
    \item \textbf{SFT Task Prompts:} Fig.~\ref{fig:sft} defines SFT tasks for ID mapping, next-item prediction, and AIGC instruction generation.
    \item \textbf{GRPO/RL Task Prompts:} Fig.~\ref{fig:grpo} defines preference prediction and instruction generation.
    \item \textbf{CID2CID/CID2INS Reasoning:} Fig.~\ref{fig:reason}  shows our reasoning process.
    \item \textbf{Novelty/Aesthetic Judging:} Fig.~\ref{fig:overall_judging} illustrates the evaluation of novelty and aesthetics for image (a) and video (b) generation.
\end{itemize}

\newpage

\clearpage

\end{document}